\documentclass{article}

\usepackage{arxiv}

\usepackage[utf8]{inputenc} 
\usepackage[T1]{fontenc}    
\usepackage{hyperref}       
\usepackage{url}            
\usepackage{booktabs}       
\usepackage{amsfonts}       
\usepackage{nicefrac}       
\usepackage{microtype}      
\usepackage{lipsum}
\usepackage{bm} 
\usepackage{graphicx}
\graphicspath{ {./images/} }
\usepackage{float}
\usepackage{natbib}
\usepackage{amsmath}
\usepackage{multirow}
\usepackage{subfigure} 

\title{Transfer learning for nonparametric Bayesian networks}

\author{
 Rafael Sojo \\
  Aingura IIoT, Paseo Mikeletegui 43, 20009 Donostia-San Sebastián, Gipuzkoa, Spain \\
  Universidad Politécnica de Madrid, Departamento de Inteligencia Artificial, 28660 Boadilla del Monte, Madrid, Spain \\
  \texttt{rafael.sojo@alumnos.upm.es} 
  \And
  Pedro Larrañaga \\
  Universidad Politécnica de Madrid, Departamento de Inteligencia Artificial, 28660 Boadilla del Monte, Madrid, Spain \\
  \texttt{pedro.larranaga@fi.upm.es} 
  \And
 Concha Bielza \\
  Universidad Politécnica de Madrid, Departamento de Inteligencia Artificial, 28660 Boadilla del Monte, Madrid, Spain \\
  \texttt{mcbielza@fi.upm.es} 
}

\begin{document}
\maketitle
\vspace{0.5cm}
\begin{abstract}
This paper introduces two transfer learning methodologies for estimating nonparametric Bayesian networks under scarce data. We propose two algorithms, a constraint-based structure learning method, called PC-stable-transfer learning (PCS-TL), and a score-based method, called hill climbing transfer learning (HC-TL). We also define particular metrics to tackle the negative transfer problem in each of them, a situation in which transfer learning has a negative impact on the model's performance. Then, for the parameters, we propose a log-linear pooling approach. For the evaluation, we learn kernel density estimation Bayesian networks, a type of nonparametric Bayesian network, and compare their transfer learning performance with the models alone. To do so, we sample data from small, medium and large-sized synthetic networks and datasets from the UCI Machine Learning repository.  Then, we add noise and modifications to these datasets to test their ability to avoid negative transfer. To conclude, we perform a Friedman test with a Bergmann-Hommel \textit{post-hoc} analysis to show statistical proof of the enhanced experimental behavior of our methods. Thus, PCS-TL and HC-TL demonstrate to be reliable algorithms for improving the learning performance of a nonparametric Bayesian network with scarce data, which in real industrial environments implies a reduce in the required time to deploy the network.
\end{abstract}


\section{Introduction}
Addressing data scarcity is an important challenge for machine learning models. By data scarcity, we refer to a problem for which there is not enough information to learn an accurate model. To address this problem, transfer learning stands out as a well-known solution for many real-world applications. It consists of improving the learning performance of a certain task by leveraging data from similar domains. By domain, we refer to a problem with a certain variable space and its corresponding marginal distributions \citep{pan_survey_2010}. By task, we refer to the learning objective (classification, clustering, regression, etc.). Similar domains are expected to have comparable behaviors and a common variable subspace, otherwise the transfer learning model could suffer negative transfer problems. Negative transfer refers to the worsening of the target's learning performance due to very dissimilar sources \citep{zhang23_negative_transfer}, a well-known problem in transfer learning applications.
Therefore, in transfer learning applications, there is a target task with scarce data, and a set of sources from similar domains with sufficient data. Some closely related terms, such as domain adaptation or multitask learning, are often presented interchangeably with transfer learning. In domain adaptation, the source domain is adapted to perform well in the target domain \citep{kouw2021-review_domainAdaptation, azarkesht_instanceDA_2022, zhou_covariance_shift_DA_2021}. Multitask learning, meanwhile, works on each task simultaneously, improving both target and source tasks by borrowing information across related domains \citep{niculescu-migzil_multitask2012, oyen_mtl-2015, benikhlef-multitask2021}. In standard transfer learning applications, the focus is exclusively placed on the target task performance. According to this, the survey \cite{pan_survey_2010} classifies these applications into three groups: inductive, if at least the target labels are available; transductive, if only the source labels are available; and unsupervised, if neither of these is available. 

In our case, we propose two unsupervised transfer learning methodologies for learning Bayesian networks, a type of probabilistic graphical model (PGM). PGMs are a field of machine learning that uses graph-based representations to encode complex distributions  \citep{koller2009probabilistic}. There is extensive literature concerning Bayesian network transfer learning for discrete data. However, it is a field that remains widely unexplored for continuous variables. There are two main methods for learning the structure of a Bayesian network: constraint-based and score-based algorithms. One of the best-known constraint-based techniques is the Peter-Clark (PC) algorithm \citep{pc-algorithm}, which evolved from the earlier SGS algorithm \citep{sgs-algorithm}. In our case, we use the more recent version, PC-stable \citep{PC-Stable}, as the basis of our constraint-based transfer learning method, named PC-stable transfer learning (PCS-TL). Regarding score-based techniques, the K2 algorithm \citep{cooper1992_K2} is one of the earliest approaches. As in PCS-TL, we use the more recent version, hill climbing (HC) \citep{hc} with tabu search \citep{glover1993tabu_search}, as the basis of our score-based transfer learning method, named hill climbing transfer learning (HC-TL). In both cases, PC-stable and HC remove the node order dependency of the original PC and K2, respectively.
We refer to the recent survey of \cite{kitson_survey_2023} for further details about Bayesian network structure learning. 

Having enough instances to learn an accurate Bayesian network is crucial. By accurate, we refer to the network that best describes the joint probability distribution (JPD) of the target's domain. Some works, such as \cite{sample_complexity1996} or \cite{dasgupta_sample_1997}, study the number of instances (sample complexity) needed for learning an accurate discrete Bayesian network from data. The former provides an upper bound for the structure learning score, whereas the latter fixes the structure and studies the sample complexity for the parameter estimation. Both studies provide sample bounds rather than exact numbers, and in both cases, the authors mention that the sample complexity depends on the number of variables and parents allowed. Therefore, increasing the number of variables can require several hundred extra instances to achieve the same level of accuracy. For instance, in structural health monitoring tasks \citep{shm} or health care applications \citep{healthcare} with dozens of variables, collecting thousands of truly representative samples for a model can take weeks or months. Consequently, our two transfer learning methods provide a reliable way of reducing the deployment time of nonparametric Bayesian networks for different domains.

The paper is organized as follows. Section \ref{sec:related_works} presents related works. Section \ref{sec:back} provides an overview of the fundamental concepts of Bayesian networks. Section \ref{sec:methodology} introduces both PCS-TL and HC-TL algorithms. Section \ref{sec:experiments} discusses the experimental results. Section \ref{sec:conclusion} concludes the paper and provides potential future research directions.

\section{Related Work}
\label{sec:related_works}

In this section, we present related works on transfer learning. These applications are divided into three groups according to the type of data: discrete, Gaussian and nonparametric. For discrete data, our focus is on Bayesian networks, and we differentiate between works performing transfer learning for structures, parameters, or both. There is limited literature on Bayesian network transfer learning for Gaussian-distributed and nonparametric data. Therefore, we have included other machine learning models.  
Finally, we summarized our contributions based on the literature review.

For discrete data, the work in \cite{pctl} stands as one of the most cited Bayesian network transfer learning approaches. They introduce the PC-TL algorithm for the structure estimation, and the distance-based linear pool (DBLP) and local linear pool (LoLP) for the fusion of parameters. PC-TL is a constraint-based algorithm, where the CI tests of the target and the most similar source task are combined according to a target confidence parameter. Here, the most similar source is selected with respect to a similarity metric. For DBLP and LoLP, they use a linear pooling \citep{linearpool} approach. In \cite{TNBN_2015}, a similar approach is proposed for temporal node Bayesian networks, while in \cite{yan_operational_2023}, the similarity metric of PC-TL is substituted by a majority vote mechanism. In this approach, the results of the CI test of multiple sources are considered instead of just the most similar. Later, \cite{yan_bayesian_2024} proposed a genetic algorithm for the structure learning process of the network. For the parameters, they extended the work of \cite{PYuan19_parameters} by removing the expert knowledge dependency. In \cite{PYuan19_parameters}, a linear pooling approach is proposed for parameter transfer learning, where the structure of the networks is given and the opinion of experts is needed for the similarity between the target and the sources. In both cases, they tested their transfer learning applications in an industrial fused magnesia smelting process. Similarly, \cite{hou_bearing_2022} proposes a varying coefficient for the parameter transfer learning of an industrial bearing fault diagnosis problem. This coefficient modulates the degree to which the parameters of the target and the sources are combined, as in the previous case, in a linear pooling approach. If the target is considered to have enough information, only its parameters are used.

For continuous Gaussian data, \cite{karbalayghareh_optimal_2018} proposes a Bayesian transfer learning method for classification problems. They use Wishart distributions as a bridge to transfer information between domains in the calculation of the target's prior probability. This improves the target's posterior and thus the classification performance. Similarly, the Bayesian network parameter learning work in \cite{zhou_when_2016} defines a "transfer prior" to improve the target's posterior. The source priors are combined based on their relatedness with the target to reduce the effect of negative transfer. The robustness against irrelevant sources is later studied. In \cite{wu_bayesian_2023}, transfer learning is studied from a Bayesian perspective, where a risk function evaluates the effect of negative transfer quantitatively. The parameters are learned by minimizing this function. This effect is also analyzed mathematically throughout the paper. 

In \cite{wu_bayesian_2023}, it is proposed a nonparametric transfer learning approach for using Dirichlet processes \citep{dirichletprocess} in a discrete, supervised, classification problem. There are other nonparametric proposals, such as: \cite{papez_transferring_2022}, for Gaussian process regression \citep{gaussianprocess2006}; \cite{alotaibi_bayesian_2022}, for object tracking under unknown conditions with Gaussian mixtures and Dirichlet process mixtures; and \cite{wang_distribution_2021} for distribution inference in stationary data streams through instance-based transfer learning. For Bayesian network-related approaches with nonparametric data, the only work somewhat related that could be found is that of \cite{zeng_dynamic_2023}. They introduce a feature learning method based on dynamic Bayesian networks for remaining useful life estimation. Here, the dynamic Bayesian network is constructed with a Markov blanket discovery algorithm. Then, the strengths of the connections with the Markov blanket of the sources and the strengths of the target are weighted to obtain a merged Markov blanket. The idea of combining sources through a weighted sum, or linear pooling, is present in many of the works listed previously.
However, it can be done for nonparametric probability density functions (PDFs) with similar results. In \cite{chen_multiple_2024}, a multiple kernel-based (MK) kernel density estimation (KDE) \citep{scott2015multivariate} is proposed. The MK-KDE model is constructed as a weighted sum of KDEs, demonstrating enhanced performance for estimating complex and multimodal PDFs. On the other hand, the work of \cite{koliander_fusion_2022} reviews the different ways in which a set of PDFs can be combined. It also lists numerous ways of estimating the weights of those PDFs.

Both PCS-TL and HC-TL methodologies are constructed based on these ideas. For PCS-TL, we use a linear pooling of nonparametric CI tests. Specifically, we use the randomized conditional correlation test (RCoT) \citep{rcot_2019}. For HC-TL, we propose a novel transfer learning score that combines nonparametric PDFs and a risk metric to avoid the effect of negative transfer. Along with that, the source weights in both transfer learning algorithms consider the divergences with the target to avoid consulting unrelated domains. In our case, we use KDE Bayesian networks (KDEBNs) \citep{KDEBN} as the nonparametric models. For the parameters, we aggregate the target with the sources through a log-linear pooling approach \citep{logpooling}. To the best of our knowledge, this is the first work proposing transfer learning techniques for nonparametric Bayesian networks. As a summary, Table \ref{tab:bns_sumary} classifies the Bayesian network transfer learning related works reviewed and their characteristics. 
Similarly, Table \ref{tab:no-bns_sumary} classifies the non-Bayesian network-related works.

\begin{table}[!ht]
  \centering
  \begin{tabular}{llccccr}
    \toprule
    \textbf{Data} & \textbf{Comments} & \multicolumn{2}{c}{\textbf{Structure}} & \textbf{Parameters} & \textbf{Negative} & \textbf{Reference} \\
                 &                    & \textbf{Constraint} & \textbf{Score}   &                     & \textbf{transfer} &                     \\
    \midrule
    Discrete      & PC-TL with DBLP and LoLP              & $\times$ &                   & $\times$ &           & \cite{pctl} \\
    Discrete      & Temporal node Bayesian networks       & $\times$ &                   & $\times$ &           & \cite{TNBN_2015} \\
    Discrete      & PC-TL with majority vote              & $\times$ &                   & $\times$ &           & \cite{yan_operational_2023} \\
    Discrete      & Genetic algorithm for the structures  &          & $\times$          & $\times$ &           & \cite{yan_bayesian_2024} \\
    Discrete      & Parameter transfer learning           &          &                   & $\times$ &           & \cite{PYuan19_parameters} \\
    Discrete      & Varying coefficient transfer learning &          &                   & $\times$ &           & \cite{hou_bearing_2022} \\
    Gaussian      & Parameter transfer learning           &          &                   & $\times$ & $\times$  & \cite{zhou_when_2016} \\
    Nonparametric & Dynamic Bayesian networks             & $\times$ &                   &          &           & \cite{zeng_dynamic_2023} \\
    Nonparametric & PCS-TL and HC-TL                      & $\times$ & $\times$          & $\times$ & $\times$  & Our work \\
    \bottomrule
    \end{tabular}
\caption{Classification of Bayesian network transfer learning related works}\label{tab:bns_sumary}
\end{table}

\begin{table}[!ht]
  \centering
  \begin{tabular}{llcr}
    \toprule
    \textbf{Data} & \textbf{Comments} & \textbf{Negative transfer} &\textbf{Reference}  \\
    \midrule
    Gaussian      & Bayesian modeling                                &          & \cite{karbalayghareh_optimal_2018}\\
    Gaussian      & Bayesian modeling                                & $\times$ & \cite{wu_bayesian_2023} \\
    Discrete      & Nonparametric modeling                           & $\times$ & \cite{wu_bayesian_2023}\\
    Nonparametric & Gaussian process regression                      &          & \cite{papez_transferring_2022} \\
    Nonparametric & Custering &          & \cite{alotaibi_bayesian_2022} \\
    Nonparametric & Distribution estimation                          &          & \cite{wang_distribution_2021} \\
    Nonparametric & Distribution estimation                          &          & \cite{chen_multiple_2024} \\
    Nonparametric & Fusion of distributions                          &          & \cite{koliander_fusion_2022} \\
     \bottomrule
    \end{tabular}
\caption{Classification non-Bayesian network transfer learning related works}\label{tab:no-bns_sumary}
\end{table}

\section{Bayesian networks}
\label{sec:back}
Let $\mathcal{B} = (\mathcal{G}, \bm{\theta})$, be a Bayesian network model that comprises a direct acyclic graph (DAG) $\mathcal{G}$ and a set of parameters $\bm{\theta}$, for which $\mathcal{G} = (V, A)$ is defined as a set of nodes $V$ and a set of arcs $A \subseteq V \times V$ between pairs of these nodes. For each $\mathcal{B}$ we have a dataset $\mathcal{D}$ with $n$ different variables $\mathbf{X} = (X_1, ..., X_n)$ and $N$ instances, i.e., $\mathcal{D} = \{\textbf{x}^{1},...,\textbf{x}^{N} \}$.  
Each arc of $A$ in $\mathcal{G}$ represents a probabilistic dependence between variables, for instance, in $X_1 \xrightarrow{} X_2$, $X_1$ is referred to as the parent of $X_2$. Therefore, for each DAG there is a set of conditional dependences (and independence) that define the parameters $\bm{\theta}$ of the conditional probability distribution (CPD) associated to each variable. Then, the joint probability distribution (JPD) of $\textbf{X}$ can be factorized using these CPDs. For continuous variables, each CPD can be seen as a conditional PDF such that the JPD factorizes as: 
\begin{equation}
    f(\textbf{x}) = \prod_{i=1}^n\ f(x_i|\textbf{x}_{\text{Pa}(i)}) \ ,
    \label{eqn:jpd_pdf}
\end{equation}
where $\text{Pa}(i)$ denotes the parents of $X_i$. 

\subsection{Kernel density estimation Bayesian networks}
KDEBNs are defined using conditional KDE (CKDE) CPDs. The density function of a multivariate KDE is:
\begin{equation}
    f_{\text{KDE}}(\textbf{x}) = \frac{1}{N} \sum_{j=1}^N K_{\textbf{H}}(\textbf{x}-\textbf{x}^{j}) \ ,
    \label{eqn:multivariate_kde}
\end{equation}
where $\textbf{x}^{j} = (x_1^j,\dots , x_n^j)$ denotes the $j$-th training instance in $\mathcal{D}$, $\textbf{H}$ represents the $n\times n$ symmetric and positive definite bandwidth matrix and $K_\textbf{H}(\textbf{x}-\textbf{x}^{j}) = |\textbf{H}|^{-1/2} K( |\textbf{H}|^{-1/2} (\textbf{x}-\textbf{x}^{j}))$ is the scaled multivariate kernel function. 
For the calculation of conditional probabilities, let $f_{\text{KDE}}(x_i,\textbf{x}_{\text{Pa}(i)})$ denote the joint KDE model for $X_i$ and $\textbf{X}_{\text{Pa}(i)}$ and $f_{\text{KDE}}(\textbf{x}_{\text{Pa}(i)})$ the marginal KDE model for $\textbf{X}_{\text{Pa}(i)}$. By using the Bayes' theorem, the conditional distribution of $X_i$ given $\textbf{X}_{\text{Pa}(i)}$ in a CKDE CPD is:
\begin{equation}
    f_{\text{CKDE}}(x_i|\textbf{x}_{\text{Pa}(i)}) = \frac{f_{\text{KDE}}(x_i,\textbf{x}_{\text{Pa}(i)})}{f_{\text{KDE}}(\textbf{x}_{\text{Pa}(i)})} =
    \frac{\sum_{j=1}^N K_{\textbf{H}_i}\left(
    \begin{bmatrix} 
    x_i\\
    \textbf{x}_{\text{Pa}(i)}\\
    \end{bmatrix} - 
    \begin{bmatrix} 
    x_{i}^{j}\\ 
    \textbf{x}_{\text{Pa}(i)}^{j}\\ 
    \end{bmatrix}\right)}{\sum_{j=1}^N K_{\textbf{H}_{i}^{-}}
    (\textbf{x}_{\text{Pa}(i)} - \textbf{x}_{\text{Pa}(i)}^{j})} \ ,
    \label{eq:non-parm-cpd}
\end{equation}
where $\mathbf{H}_i$ and $\mathbf{H}_{i}^{-}$ are the joint and marginal bandwidth matrices for $f_{\text{KDE}}(x_i,\textbf{x}_{\text{Pa}(i)})$ and $f_{\text{KDE}}(\textbf{x}_{\text{Pa}(i)})$, respectively.

\subsection{Parameter learning}
\label{sec:structure_learning}

The main parameters of a CKDE node are the bandwidth matrices $\textbf{H}_i$ and $\textbf{H}_i^-$. A common approach for the estimation of the bandwidth matrix is the normal reference rule extended to the multivariate case \citep{normal-reference-rule}:
\begin{equation}
    \hat{\mathbf{H}}_i = \left(\frac{4}{n+2}\right)^{2/(n+4)} \bm{\hat{\Sigma}} N^{-2/(n+4)} \label{eq:normal_rule} ,
\end{equation}
where $\bm{\hat{\Sigma}}$ is the sample covariance matrix of $X_i$ and $\textbf{X}_{\text{Pa}(i)}$. 
The normal reference rule provides a closed-form solution for minimizing the asymptotic approximation of the mean integrated squared error (AMISE). The AMISE is an approximation to the mean integrated squared error (MISE) when $N \xrightarrow{} \infty$, which is given by the expectation of the integrated squared error (ISE):
\begin{equation}
    \text{MISE}\{\hat{f}\} = \mathbb{E} \left[ \text{ISE}\{\hat{f}\} \right] = \mathbb{E} \left[ \int_{\mathbb{R}^n} \left(\hat{f}(\textbf{x}) - f(\textbf{x})\right)^2 d\textbf{x}\right]
\end{equation}
where $f(\textbf{x})$ is the true density function and $\hat{f}(\textbf{x})$ is the estimate.
\subsection{PC-stable}
Constraint-based algorithms start with a complete undirected graph and comprehend three phases: an adjacency phase; a v-structure phase; and a final orientation propagation phase, where the remaining edges are oriented as possible. 
The adjacency phase performs CI tests over pairs of $X$ and $Y$ variables to asses whether the edge $X - Y$ should be removed or not. This step is done for conditioning sets of nodes $\textbf{Z}$ with increasing size (Sepsets). Under the assumption of faithfulness and causal sufficiency, the search space is reduced to nodes adjacent to $X$ and $Y$. In PC, the removal of the edge  $X-Y$, if needed, is done after the CI test. In contrast, PC-stable removes the edges after finishing all the CI tests for that Sepset size, thus eliminating the node order dependency. 

Then, the v-structure phase re-uses Sepsets identified during the adjacency phase to assess unshielded triples of $X-Y-Z$. The term unshielded refers to the fact that there is no edge connecting $X$ and $Z$. The collider $X\xrightarrow{}Y\xleftarrow{}Z$ is oriented if $X$ and $Z$ are conditionally independent given $Y$. In PC-stable, they adopted the approach proposed in \cite{Conservative-pc}  with weaker restrictions. Here, the v-structure is selected if the majority of Septets from $X$ and $Z$ in the triple $X-Y-Z$ do not contain $Y$.
 
Finally, the orientation propagation phase follows the "Meek rules" \citep{meek_rules}. The result of this process is a partially directed graph (PDAG). Therefore, to convert the PDAG into a DAG, we refer to the algorithm presented in \cite{pdag2dag}.

\subsection{Hill climbing}
Contrary to PC-stable, HC starts with an empty graph. Then, different arc additions, arc deletions, and arc flips are performed and evaluated according to a score function. For this task, several scores rely on the computation of the log-likelihood of a dataset $\mathcal{D}$ given a DAG $\mathcal{G}$ and some parameters $\bm{\theta}$. Assuming independent identically distributed samples in $\mathcal{D}$, let the log-likelihood $\mathcal{L}$ of $\mathcal{D}$ given $\mathcal{G}$ and $\bm{\theta}$ be:
\begin{equation}
    \mathcal{L}(\mathcal{D} | \mathcal{G},\bm{\theta}) = \sum_{i=1}^{n} \mathcal{L}( \mathcal{D} | \mathcal{G},\bm{\theta}_i) =\sum_{i=1}^{n} \sum_{j=1}^N \log f(x_{i}^{j}|\textbf{x}_{\text{Pa}(i)}^{j})
    \label{eq:logl}
\end{equation}
where $\bm{\theta}_i$ corresponds to the set of parameters associated to the CPD of node $i$. 
However, as explained in \cite{SPBN}, any score including the log-likelihood of the training data would be inappropriate for the KDE, since the data constitutes part of the model (see Equation \eqref{eqn:multivariate_kde}). Therefore,  we refer to their $k$-fold cross-validated log-likelihood function $\mathcal{S}_{\text{CV}}^k$:
\begin{align}
    \label{eq:sk_cv}
    \mathcal{S}_{\text{CV}}^k(\mathcal{D},\mathcal{G}) = \sum_{m=1}^k \sum_{i=1}^{n} \mathcal{L}( \mathcal{D}^{\mathcal{I}^m} | \mathcal{G},\bm{\theta}_i^{\mathcal{I}^{-m}_{\text{train}}})
\end{align}
where $\mathcal{L}( \mathcal{D}^{\mathcal{I}^m} | \mathcal{G},\bm{\theta}_i^{\mathcal{I}^{-m}_{\text{train}}})$ denotes the log-likelihood of the $m$-th fold data, given a graph $\mathcal{G}$ with $\bm{\theta}_i^{\mathcal{I}^{-m}_{\text{train}}}$ parameters for node $i$. These parameters are estimated for the training fold data $\mathcal{D}^{\mathcal{I}^{-m}_{\text{train}}}$, considering that $\mathcal{I} = \{\mathcal{I}^1,\dots, \mathcal{I}^k\}$ denotes $k$ disjoint sets of indices and $\mathcal{I}^{-m}_{\text{train}} = \bigcup_{l=1: l\neq m}^k \mathcal{I}^l$ are the indices not in the $m$-th fold.

\section{Transfer learning for nonparametric Bayesian networks}
\label{sec:methodology}
\begin{figure}[H]
    \centering
    \includegraphics[width=1\linewidth]{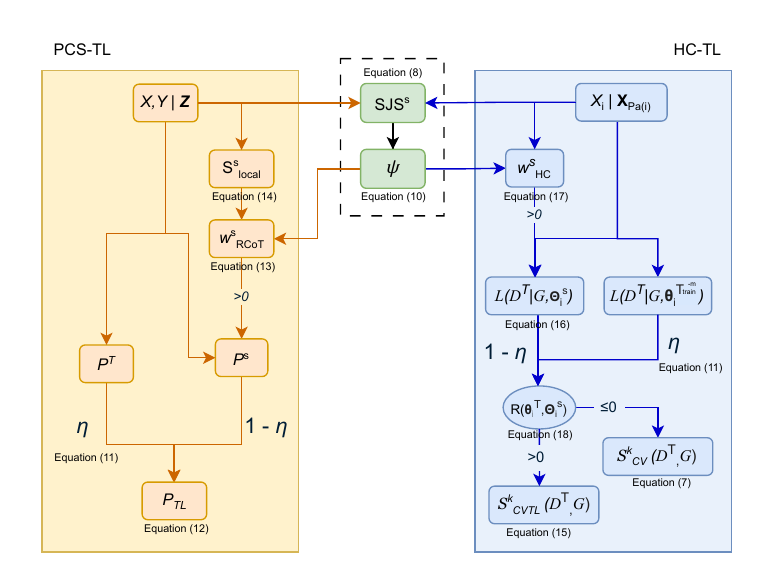}
    \caption{Flowchart of PCS-TL (in orange) and HC-TL (in blue). In green, functions that are shared by both processes.}
    \label{fig:flowchart_transfer}
\end{figure}

As explained in the introduction, our work proposes a nonparametric Bayesian network transfer learning framework for constraint-based and score-based structure learning algorithms. We will linearly combine the opinions of different sources with the target to enhance its learning performance under scarce data. Then, to mitigate the effect of negative transfer, we will measure the similarity between the sources and the target. Here, very dissimilar sources are considered outliers and set aside. In both algorithms, the opinion of the remaining sources will be averaged according to their relatedness. However, both learning algorithms have their particular considerations for reducing negative transfer. Finally, we assume that the source tasks are learned with sufficient data. Therefore, they are suitable for combination with the target. To do so, we propose the target trust factor $\eta$. In Figure \ref{fig:flowchart_transfer}, we illustrate the flowchart of the transfer learning framework for PCS-TL and HC-TL. In the following subsections, we will break down the particular metrics we propose for each algorithm, as well as the common ones to both (shown in green).

\subsection{Jensen-Shannon and target trust factor}
Let $\mathcal{D}^T$ be the target dataset of a particular domain, with $n$ different variables $\textbf{X}=(X_1,\dots,X_n)$ and $N^T$ instances. 
Likewise, let $\bm{\mathcal{D}}^S = \{\mathcal{D}^s\}_{s=1}^S$, $S\geq1$, be the set of source datasets from similar domains, with the same $n$  variables and corresponding $\textbf{N}^S=\{N^1, \dots, N^S\}$ instances.
To compute the similarity between the target (index $T$) and the sources (index $s$), we propose the sum of Jensen-Shannon (SJS) divergence:
\begin{equation}
    \text{SJS}^s(\textbf{X}) = \sum_{X \in \textbf{X}} \text{JS}(f_{\text{KDE}}^T(X) \| f_{\text{KDE}}^s(X)) \ \ \text{ for }s=1,\dots,S
    \label{eq:sjs_diver}
\end{equation}
where $\text{JS}(f_{\text{KDE}}^T(X) \| f_{\text{KDE}}^s(X))$ denotes the Jensen-Shannon (JS) divergence between the probability distributions $f_{\text{KDE}}^T(X)$ and $f_{\text{KDE}}^s(X)$ for the target and source $s$, respectively. In our case, these distributions are estimated using the KDE approach described in Equation \eqref{eqn:multivariate_kde}, but any other suitable method can be employed. The JS divergence is defined as follows.
\begin{equation}
  \mathrm{JS}(P \| Q) = \frac{1}{2} \mathrm{KL}(P \| M) + \frac{1}{2} \mathrm{KL}(Q \| M)  
\end{equation}
where KL is the Kullback-Leibler divergence, $P$ and $Q$ are probability distributions and  $M = 0.5(P + Q)$. We use JS because it is a smoothed, symmetrized version of KL.
The SJS divergence is computed for every pair of target and source $s$. If there are $S$ sources, they will be computed $S$ SJS divergences. 
To assess whether the domain $\mathbf{X}$ of a source $s$ is related to the target domain, the function $\psi$ is applied to each SJS value. This function determines if a value is considered too large (an outlier) compared to other sources. The larger the SJS value, the more dissimilar the domains are. If the SJS value is not considered an outlier, $\psi$ returns the inverse of the original value; otherwise, it returns 0. We return the inverse to give a higher rank to those sources that are more similar to the target, as the output of $\psi$ is subsequently used to compute source weights in PCS-TL and HC-TL (see Equation \eqref{eq:w_rcot} and Equation \eqref{eq:weights_hc}, respectively). It is defined as:
\begin{equation}
    \psi(u^s) = 
    \begin{cases}
        \frac{1}{u^s} & \text{if} \hspace{0.5cm} u^s \leq Q_3(\textbf{u})+1.5 \cdot\text{IQR}(\textbf{u})\\
        0   &\text{otherwise}
    \end{cases}
    \label{eq:psi_sjs}
\end{equation}
where $u^s$ is the $s$-th SJS value of $\textbf{u}=(u^1,\dots,u^S)$, $Q_3$ is a function that returns the third percentile of a list of values, and IQR is a function that returns the interquartile range. In this case, for the list of SJS divergences $\textbf{u}$. This way, we avoid using very dissimilar sources in a particular domain, but allow the same sources to be considered for another set of variables during the learning process. 

On the other hand, the knowledge of the sources is combined with the knowledge of the target through the target trust factor $\eta$. This factor allows the target's scarce information to be complemented by the sources (see Sections \ref{sec:constrained} and \ref{sec:scored}). Since the sources are considered to be learned with sufficient data, it is constructed as the number of target instances over the average source instances. Let $\eta$ be defined as:
\begin{equation}
    \eta=\frac{N^T}{\frac{1}{S^{+}}\sum_{s\in \textbf{S}^+}N^{s}}, \hspace{0.7cm} \eta\in(0,1]. 
    \label{eq:etafactor}
\end{equation}
where $\textbf{S}^+$ denotes the set of indices from $\textbf{N}^S$, with size $S^+$, for which $\psi(u^s)>0$. Note that $\eta$ is restricted to values $0<\eta \leq 1$. Thus, if the target's dataset $\mathcal{D}^T$ is sufficiently large, it allows the network's learning to gradually shift into a target-only task, instead of a transfer learning task.

\subsection{PC-stable transfer learning}
\label{sec:constrained}
As we introduced previously, we use PC-stable as the base structure learning algorithm for PCS-TL. To correctly evaluate the relationship between pairs of continuous nonparametric variables, we require an accurate nonparametric CI test. In our case, we use RCoT for this task. The RCoT CI test uses random Fourier features \citep{Rahimi2007RandomFF} to approximate the results of the kernel conditional independence test \citep{kcit} several orders of magnitude faster with nearly identical accuracy. To learn the structure of the target considering the opinion of the auxiliary sources, we use a linear pooling operation to combine the $p$-values of the CI tests of the target and the sources. Let $P_{\text{TL}}(X,Y|\textbf{Z})$ be the combined, transfer learning, $p$-value for the pair of variables $X$ and $Y$ given a subset $\textbf{Z}$. If $P_{\text{TL}}(X,Y|\textbf{Z})$ is smaller than a significance level $\alpha$, the edge between $X$ and $Y$ is removed. 
It is defined as:
\begin{align}
   P_{\text{TL}}(X,Y|\textbf{Z}) &= \eta \ P^T(X,Y|\textbf{Z}) + (1-\eta)\sum_{s=1}^{S} w^s_{\text{RCoT}} \cdot P^s(X,Y|\textbf{Z}) 
   \label{eq:pval_transfer}
\end{align}
where
\begin{align}
   w^s_{\text{RCoT}} &= \frac{\psi(\text{SJS}^s(X,Y,\textbf{Z}))\cdot S^s_{\text{local}}}{\sum_{s=1}^S \psi(\text{SJS}^s(X,Y,\textbf{Z}))\cdot S^s_{\text{local}}}  
   \ \ \text{ for }s=1,\dots,S
   \label{eq:w_rcot}
\end{align}
where $P^T(X,Y|\textbf{Z})$ is the RCoT $p$-value of the target, $P^s(X,Y|\textbf{Z})$ is the RCoT $p$-value of the source $s$  with corresponding weight $w^s_{\text{RCoT}}$ and $\eta$ is the target trust factor given given by Equation \eqref{eq:etafactor}.
To mitigate the effect of negative transfer for constraint-based algorithms, this weight is computed based on the inverse of the SJS divergence (Equation \eqref{eq:sjs_diver} and Equation \eqref{eq:psi_sjs}) and the local similarity $S^s_{\text{local}}$ \citep{pctl}. The local similarity $S^s_{\text{local}}$ between the source $s$ and the target is defined as:
\begin{equation}
    S^s_{\text{local}}(X,Y|\textbf{Z}) =
    \begin{cases}
        1 &\text{if} \hspace{0.7 cm} I^T(X,Y|\textbf{Z}) = I^s(X,Y|\textbf{Z})\\
        0.5 &\text{otherwise}
    \end{cases}
\end{equation}
where $I^T$ and $I^s$ are the CI tests of the target and the source $s$, respectively. In other words, $S^s_{\text{local}}$ equals 1 if both $p$-values are greater or minor than the significance level $\alpha$. This metric prioritizes sources with similar opinions to the target while considering the rest of the sources. Thus, the most similar sources will have a higher weighting than the rest.

\subsection{Hill climbing transfer learning}
\label{sec:scored}
For HC-TL, we use HC. To consider the opinion of the sources during the learning process, we propose the $k$-fold cross-validated transfer learning log-likelihood $\mathcal{S}_{\text{CVTL}}^k$, adapted from $\mathcal{S}_{\text{CV}}^k$ (Equation \eqref{eq:sk_cv}):
\begin{align}
    \label{eq:sk_cvtl}
    \mathcal{S}_{\text{CVTL}}^k(\mathcal{D}^T, \mathcal{G}) = \sum_{m=1}^k \mathcal{L}_{\text{TL}}( \bm{\theta}^T, \bm{\Theta}^S) =  \sum_{m=1}^k \sum_{i=1}^{n} (\eta \  \mathcal{L}( \mathcal{D}^T | \mathcal{G},{\bm{\theta}_i}^{T^{-m}_{\text{train}}}) + (1-\eta) \ \mathcal{L}( \mathcal{D}^T | \mathcal{G},{\bm{\Theta}^S_i}))
\end{align}
where ${\bm{\theta}_i}^{T^{-m}_{\text{train}}}$ denotes the $m$-th fold target training parameters of node $i$, and ${\bm{\Theta}^S_i} = \{ {\bm{\theta}^1_i}, \dots, {\bm{\theta}^S_i} \}$ denotes the set of source parameters of node $i$. Note that ${\bm{\theta}_i}^{T^{-m}_{\text{train}}}$ is obtained for the training fold data $\mathcal{D}^{T^{-m}_{\text{train}}}$, while the set ${\bm{\Theta}^S_i}$ is obtained for $\bm{\mathcal{D}}^S$.
Therefore, $\mathcal{L}( \mathcal{D}^T | \mathcal{G},{\bm{\theta}_i}^{T^{-m}_{\text{train}}})$ is the log-likelihood of the $m$-th fold target dataset $\mathcal{D}^T$ given ${\bm{\theta}_i}^{T^{-m}_{\text{train}}}$, and $\mathcal{L}( \mathcal{D}^T | \mathcal{G},{\bm{\Theta}^S_i})$ is the source log-likelihood weighted average of $\mathcal{D}^T$ given the set of parameters ${\bm{\Theta}^S_i}$. This weighted average is computed as:
\begin{align}
   \mathcal{L}( \mathcal{D}^T | \mathcal{G},{\bm{\Theta}^S_i}) &= \sum_{s=1}^{S} w^s_{\text{HC}} \cdot \mathcal{L}( \mathcal{D}^T | \mathcal{G},{\bm{\theta}^s_i})
\end{align}
where $w^s_{\text{HC}}$ is the weight of the source $s$, defined as:
\begin{align}
   w^s_{\text{HC}} &= \frac{\psi(\text{SJS}^s(X_i,\textbf{X}_{\text{Pa}(i)}))}{\sum_{s=1}^S \psi(\text{SJS}^s(X_i,\textbf{X}_{\text{Pa}(i)}))} 
   \ \ \text{ for }s=1,\dots,S
   \label{eq:weights_hc}
\end{align}
 
Additionally, we propose a risk metric to avoid the effect of negative transfer in the context of a score-based  structure learning algorithm:
\begin{equation}
    R(\bm{\theta}^T_i, \bm{\Theta}^S_i) = |\mathcal{L}( \mathcal{D}^T | \mathcal{G},\bm{\theta}^T_i)| - | \mathcal{L}( \mathcal{D}^T | \mathcal{G},\bm{\Theta}^S_i)|
    \label{eq:risk}
\end{equation}
where $|\cdot|$ refers to the absolute value of each log-likelihood. According to this, $R(\bm{\theta}^T_i, \bm{\Theta}^S_i) >0$ implies that the combined effect of sources benefits the target. In that case, $\mathcal{L}( \mathcal{D}^T | \mathcal{G},{\bm{\theta}_i}^{T^{-m}_{\text{train}}})$ and $\mathcal{L}( \mathcal{D}^T | \mathcal{G},{\bm{\Theta}^S_i})$ are combined according to the target trust factor $\eta$. However, if $R(\bm{\theta}^T_i, \bm{\Theta}^S_i) \leq 0$, the combined effect of sources is harmful for the target. In that case, $\mathcal{S}_{\text{CV}}^k(\mathcal{D}^T, \mathcal{G})$ (Equation \eqref{eq:sk_cv}) is used instead of $\mathcal{S}_{\text{CVTL}}^k(\mathcal{D}^T, \mathcal{G})$ (Equation \eqref{eq:sk_cvtl}). In other words, the sources are not considered for the learning process if their log-likelihood is greater than the target's log-likelihood in absolute value.

\subsection{Parameter transfer learning}
Once we have the structure of the network, the CPD of each node in the transfer learning KDEBN is estimated as a weighted geometric mean with a log-linear pooling of the target and sources. Since the structure of each source can be different, their parameters ($\textbf{H}_i$ and $\textbf{H}_i^-$) are calculated for the new transfer learning structure. These sources are combined following the same log-linear pooling approach. Here, the weights of each source are estimated as in Equation \eqref{eq:weights_hc}. Then, the resulting CPDs are combined with the target according to the target trust factor $\eta$ (Equation \eqref{eq:etafactor}). Thus, let the CKDE transfer learning (CKDE-TL) CPD be defined as:
\begin{align}
    f_{\text{CKDE-TL}}(x_i|\textbf{x}_{\text{Pa}(i)}) &= \ f^T_{\text{CKDE}}(x_i|\textbf{x}_{\text{Pa}(i)})^{\ \eta} \cdot  f^S_{\text{CKDE}}(x_i|\textbf{x}_{\text{Pa}(i)})^{\ 1-\eta} 
\end{align}
where 
\begin{align}
    f^S_{\text{CKDE}}(x_i|\textbf{x}_{\text{Pa}(i)}) &= \prod_{s=1}^S f^s_{\text{CKDE}}(x_i|\textbf{x}_{\text{Pa}(i)})^{ w^s_{\text{HC}}}
\end{align}
and $f^T_{\text{CKDE}}(x_i|\textbf{x}_{\text{Pa}(i)})$ is the target CKDE CPD of node $X_i$ given $\textbf{X}_{\text{Pa}(i)}$, and $f^s_{\text{CKDE}}(x_i|\textbf{x}_{\text{Pa}(i)})$ is the source $s$ CKDE CPD for the same $X_i$ and $\textbf{X}_{\text{Pa}(i)}$.

\section{Experimental results}
\label{sec:experiments}
We will evaluate the performance of PCS-TL and HC-TL methodologies through a set of experiments. First, we will sample data from synthetic networks to compare the structure of the target task alone with the target task with transfer learning (section \ref{sec:experiment_synthetic}). Second, we will use data from the UCI Machine Learning repository \citep{uci_repo} to demonstrate consistency for different datasets (section \ref{sec:experiment_uci}). In both cases, we will evaluate the log-likelihood of a test dataset with 1024 instances and measure the time taken by the algorithms. Then, we will perform a Friedman test followed by a Bergmann-Hommel \textit{post-hoc} analysis \citep{friedman_with_berg-hommel} (section \ref{sec:experiment_bergmann}). This analysis identifies pairwise statistically significant differences between the results. To this end, our models will prove to be statistically suitable for situations of target data scarcity with nonparametric Bayesian networks. All the experiments will be repeated 3 times with different sample seeds to ensure consistency. The average results will be presented with their corresponding standard deviations in a chart.  

For the KDEBNs, we will use the normal reference rule (Equation \eqref{eq:normal_rule}) and the Gaussian kernel $K(\textbf{x}) = (2\pi^{n/2})^{-1}e^{-\frac{1}{2}\textbf{x}^\text{T}\textbf{x}}$. However, any other kernel with a valid $\textbf{H}$ can be used \citep{Chacon2018,kernel_smooth}. To implement and evaluate HC-TL, we will use a modified version\footnote{\url{https://repo.hca.bsc.es/gitlab/aingura-public/pybnesian}} of the PyBNesian python package \citep{pybnesian} with a patience of $\lambda=3$ and 5 folds. Thus, the KDEBNs will be learned using the $\mathcal{S}_{\text{CV}}^k$ (Equation \eqref{eq:sk_cv}) or $\mathcal{S}_{\text{CVTL}}^k$ (Equation \eqref{eq:sk_cvtl}) score, depending on whether the network uses transfer learning or not. Likewise, we will use the modified PyBNesian for implementing and evaluating PCS-TL with a significance level $\alpha=0.05$. Both implementations have been parallelized as much as possible to optimize the time taken by the algorithms.

\subsection{Synthetic networks}
\label{sec:experiment_synthetic}
Since there are no known synthetic nonparametric Bayesian networks available online, we will sample data from:
\begin{itemize}
    \item Four small-sized semiparametric Bayesian networks (SPBNs) \citep{SPBN}, to simulate real-world scenarios with both parametric and nonparametric distributions.
    \item Two medium and large-sized Gaussian Bayesian networks (GBNs) \citep{LGBNs}, to test the performance for a higher number of variables.
\end{itemize}

\begin{figure}[!ht]
    \centering
    \subfigure[Synthetic SPBN 1.]{
        \includegraphics[width=0.16\linewidth]{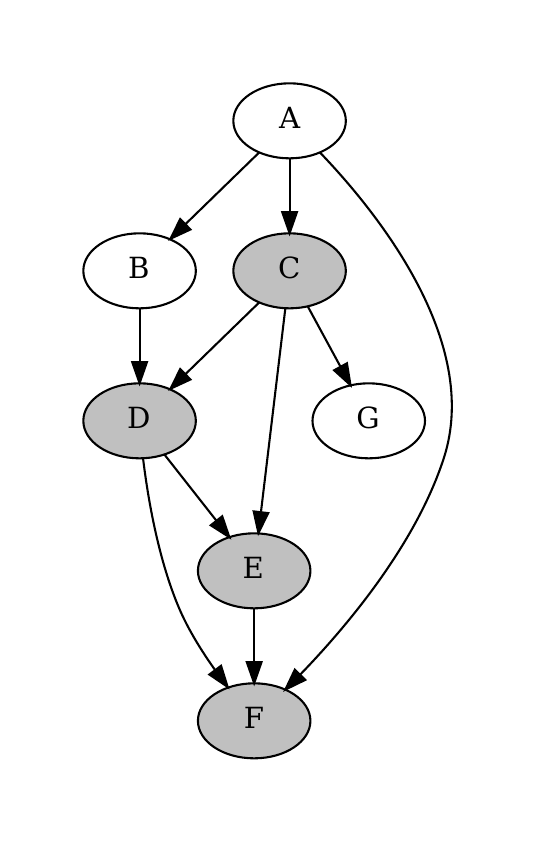}
        \label{fig:synthetic1}
    }
    \hfill
    \subfigure[Synthetic SPBN 2.]{
        \includegraphics[width=0.27\linewidth]{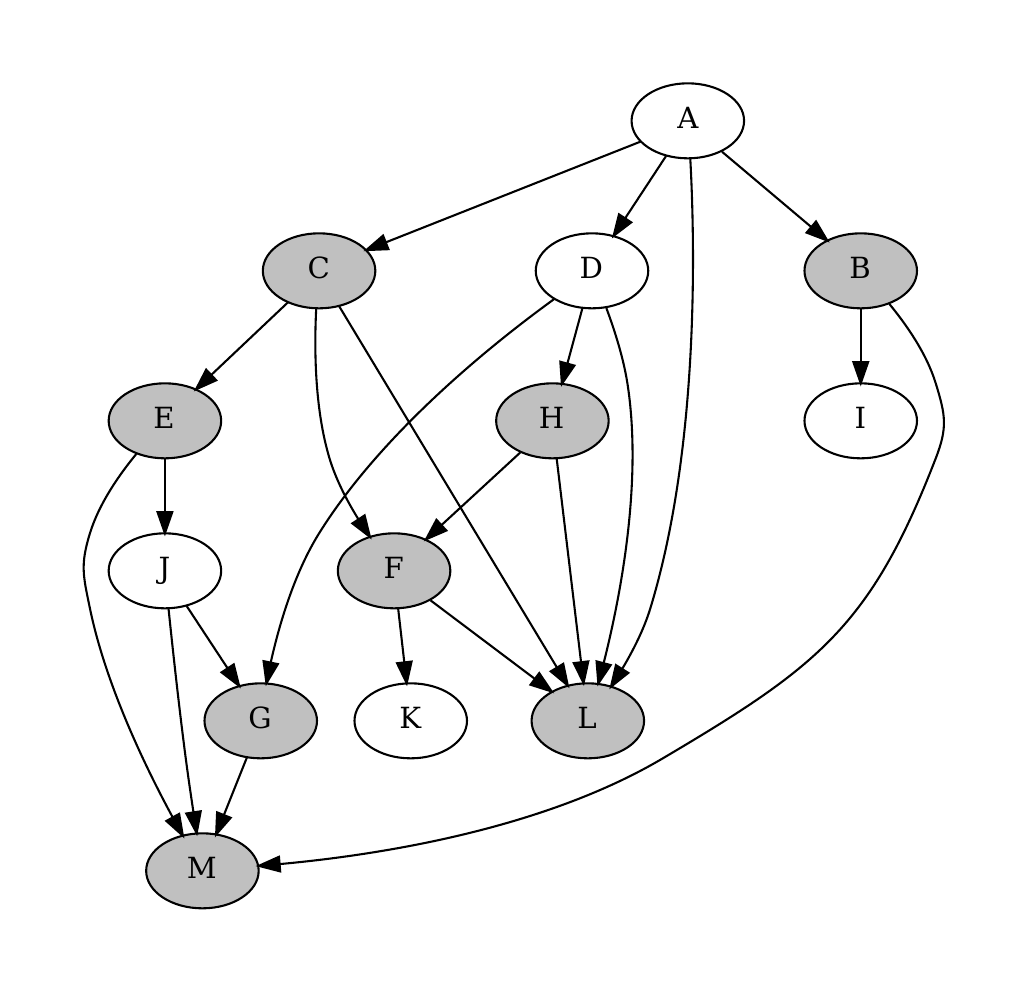}
        \label{fig:synthetic2}
    }
    \hfill
    \subfigure[Synthetic SPBN 3.]{
        \includegraphics[width=0.22\linewidth]{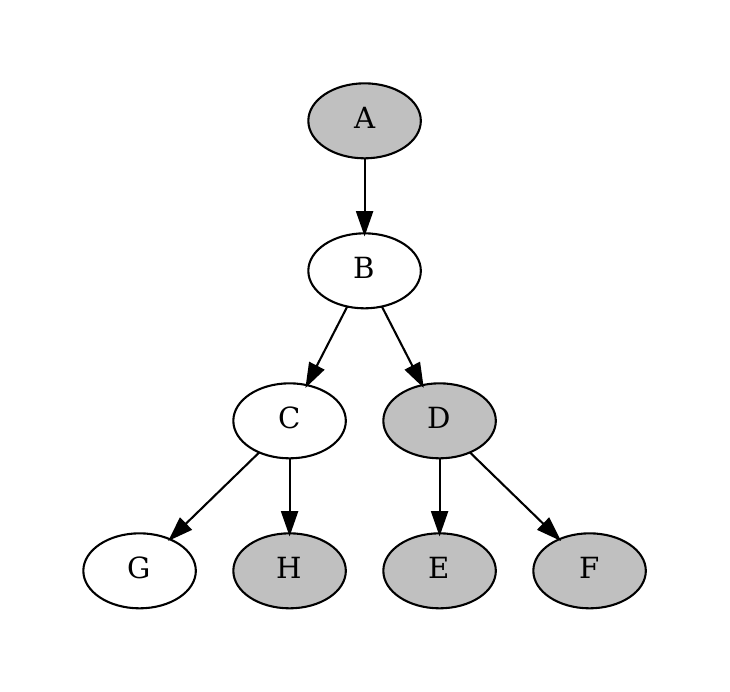}
        \label{fig:synthetic3}
    }    
    \hfill
    \subfigure[Synthetic SPBN 4.]{
        \includegraphics[width=0.27\linewidth]{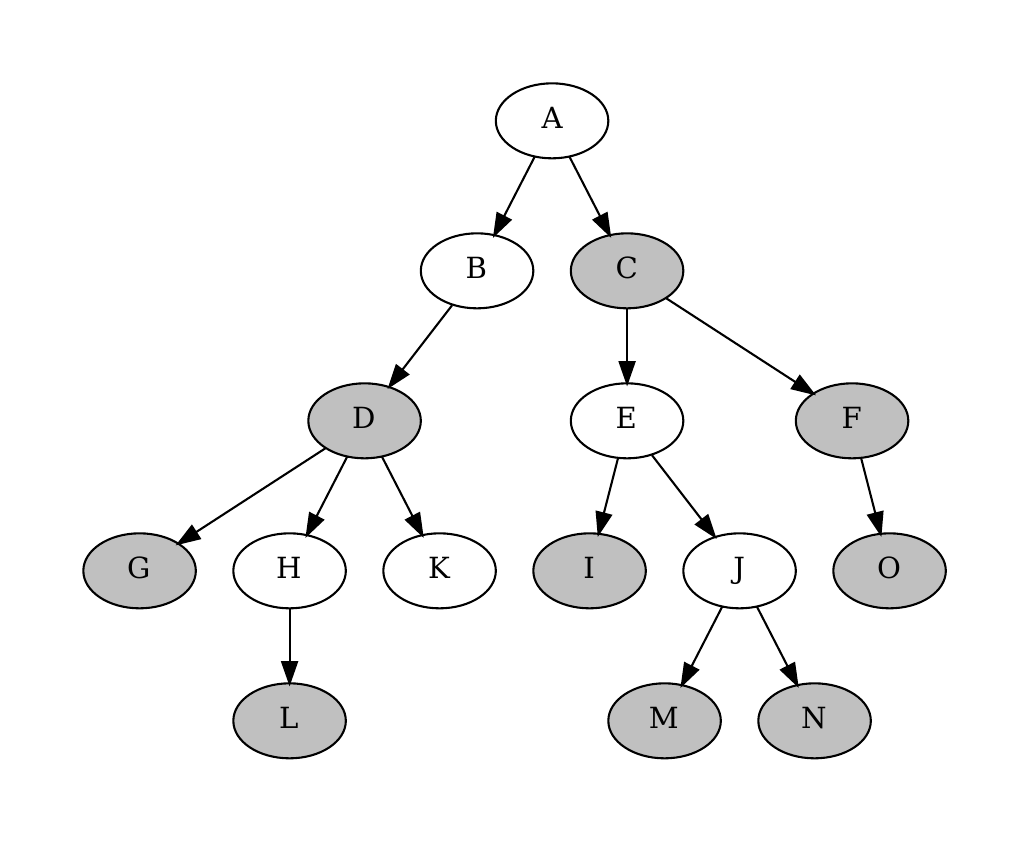}
        \label{fig:synthetic4}
    }
    \caption{Structures of the synthetic SPBNs \citep{bspbn_sojo25}.}
    \label{fig:synthetics}
\end{figure}

For the synthetic SPBNs, we will use a set of functions (see Appendix \ref{apendix:synthetic_data}) from which to sample data to construct the networks. These functions were defined by \cite{bspbn_sojo25} in a previous work. Figure \ref{fig:synthetics} illustrates their structures and Table \ref{tab:synthetic_characteristics} summarizes their main characteristics: number of nodes, number of arcs, and maximum number of parents per node (max indegree). Nodes in white correspond to Gaussian distributions, while nodes in gray are modeled through CKDE CPDs. For the synthetic GBNs, the difficulty of constructing medium and large-sized networks (which is not the scope of this work) led us to visit bnlearn \citep{bnlearn}, a well-known Bayesian network repository. From bnlearn, we will use: magic-niab, as a medium-sized network; and magic-irri, as a large-sized network. Although GBNs are only applicable to Gaussian nodes, KDEs can be used for any probability distribution. Therefore, we will use these networks for sampling new data for the target and auxiliary sources, and as a reference for measuring the structural error (see next paragraph). Table \ref{tab:bnlearn_characteristics} summarizes their main characteristics.

\begin{table}[!ht]
    \centering
    \begin{tabular}{cccc}
    \toprule
    \textbf{Model} & \textbf{Nodes} & \textbf{Arcs} & \textbf{Max indegree}  \\
    \midrule
     Synthetic SPBN 1 & 7 & 10 & 3 \\
     Synthetic SPBN 2 & 13 & 21 & 5 \\
     Synthetic SPBN 3 & 8 & 7 & 1 \\
     Synthetic SPBN 4 & 15 & 14 & 1 \\
     \bottomrule
    \end{tabular}
     \caption{Characteristics of the synthetic SPBNs.}\label{tab:synthetic_characteristics}
\end{table}

\begin{table}[!ht]
  \centering
  \begin{tabular}{lccc}
    \toprule
    \textbf{Model} & \textbf{Nodes} & \textbf{Arcs} & \textbf{Max indegree}  \\
    \midrule
     Magic-niab & 44	& 66  & 9 \\
     Magic-irri & 64	& 102 & 11 \\
     \bottomrule
    \end{tabular}
\caption{Characteristics of the bnlearn GBNs}\label{tab:bnlearn_characteristics}
\end{table}

In addition to the log-likelihood and the time taken by the algorithms, we will evaluate the error of the proposed approaches during the structure estimation. To do so, we will use the structural Hamming distance (SHD) and the density of arcs $\rho$. The SHD measures the number of arc additions, deletions, and flips required to transform one DAG (the estimated) into another (the real). The density of arcs $\rho$ refers to the total number of arcs in the model's structure. In Bayesian network transfer learning, we want to penalize empty graphs as well as reward low structural errors. However, two different structures may yield the same SHD: a transfer-learning estimate with several arcs incorrectly oriented, and a target-only estimate that is empty. In both cases, SHD does not distinguish between these qualitatively different errors. To address this situation, we propose the density-Hamming distance (DHD). DHD behaves as SHD, but incorporates the absolute difference in arc density between the true and estimated graphs. Under this framework, both empty and very complex graphs will exhibit large differences in arc densities, increasing the structural error, but only perfect estimations will still ensure a zero value. This way, we evaluate the structure and penalize empty graphs without rewarding very complex results.
DHD is defined as:
\begin{equation}
    \text{DHD} =\text{SHD} \cdot (1+|\rho_{\text{true}} - \rho_{\text{pred}}|)
\end{equation}
where the offset $+1$ ensures that only perfect estimates yield a zero value, and $\rho_{\text{real}}$ and $\rho_{\text{pred}}$ are the arc densities of the real and estimated DAGs, respectively.

The experiments should also consider the negative transfer effects. Modifying the arcs of a network can make the probability distribution of the variables significantly different from the original due to the change in the CPDs.  
To this end, we randomly modified a percentage of the arcs of the true Bayesian networks and sampled new data in each case. Then, we added random Gaussian noise to the sampled datasets. These noisy, sampled datasets from modified networks will form part of the auxiliary sources. The random Gaussian noise added has a mean of 0 and a standard deviation of 1.
The auxiliary sources that will be used, along with their respective modification percentages, are:
\begin{itemize}
    \item Two auxiliary sources with 0\% and 10\% of the arcs modified, respectively.
    \item Three auxiliary sources with  5\%, 10\% and 20\% of the arcs modified, respectively.
\end{itemize}

The rationale behind these perturbation percentages and the number of auxiliary sources is to simulate a real problem, where we have various sources (two or three) with some of them very similar to the target (0\% and  5\%), and some others less similar (10\% and 20\%). However, in a real problem, we will never use a completely different source for transfer learning, as we intend to improve the target's performance. In our case, we set this limit around 20\%. The way these are combined compares a more desirable scenario with fewer sources and differences (0\% and  10\%), and a more disadvantageous scenario with higher differences and resource needs (5\%, 10\%, and 20\%).

\begin{figure}[!htb]
    \centering
    \subfigure{
        \includegraphics[width=.98\linewidth]{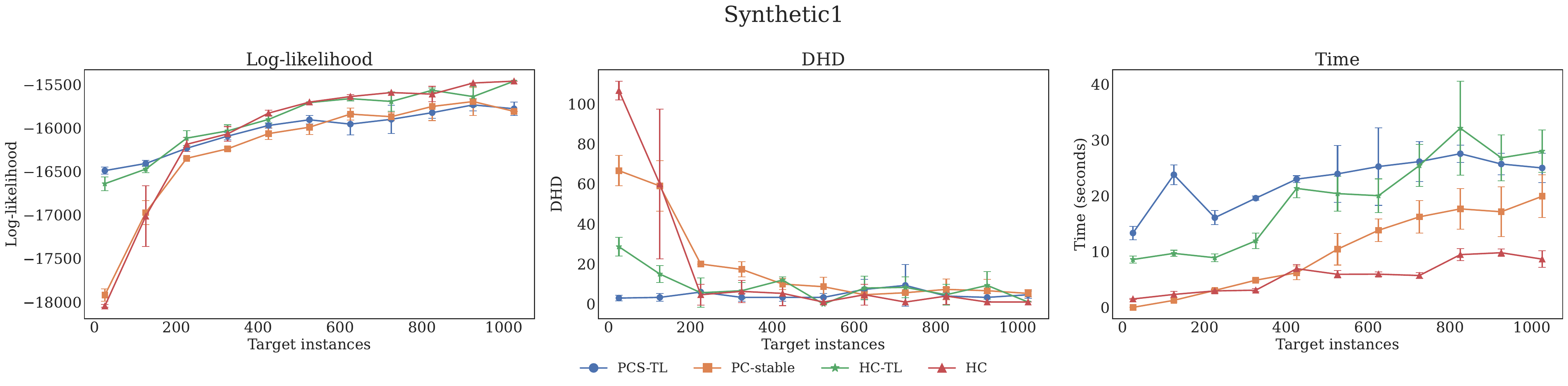}
        \label{fig:s1_010p}
    }
    \subfigure{
        \includegraphics[width=.98\linewidth]{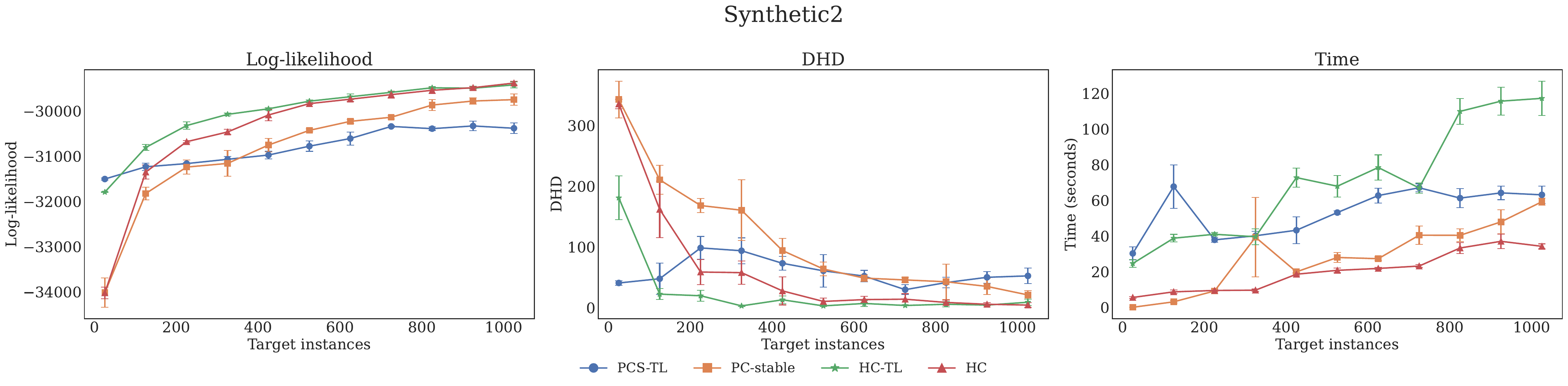}
        \label{fig:s2_010p}
    }
    \subfigure{
        \includegraphics[width=.98\linewidth]{ 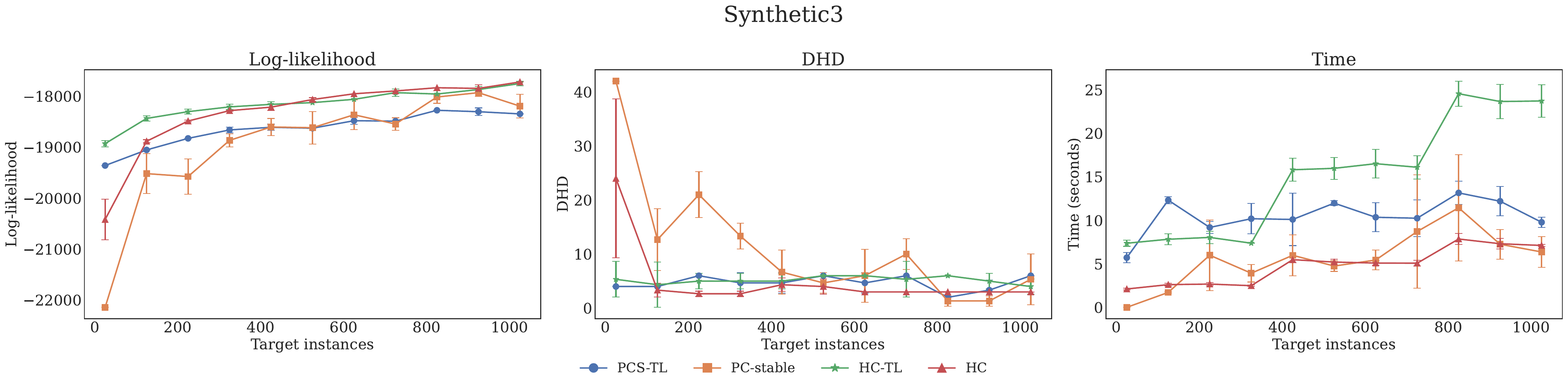}
        \label{fig:s3_010p}
    }    
    \subfigure{
        \includegraphics[width=.98\linewidth]{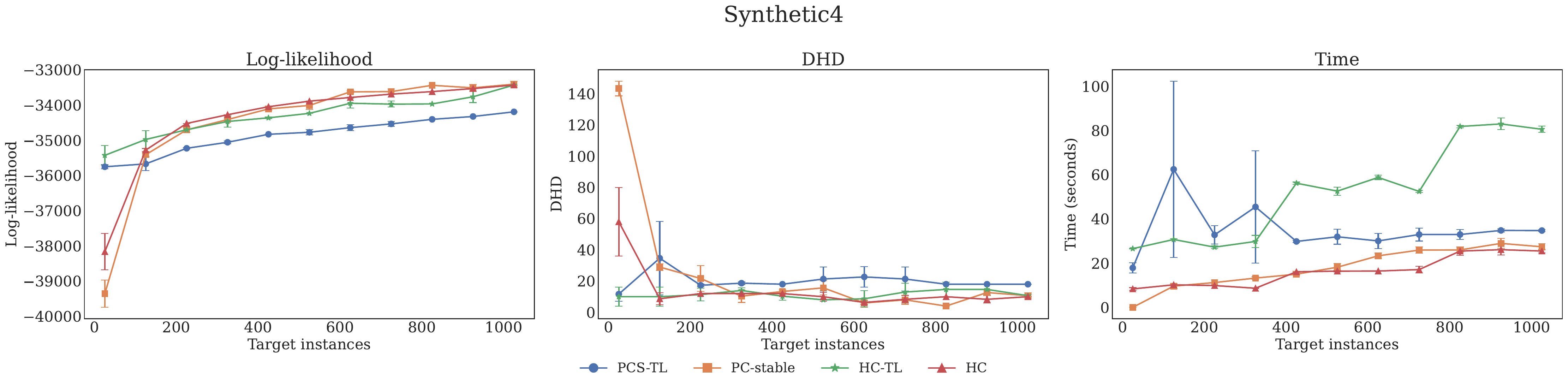}
        \label{fig:s4_010p}
    }
    \caption{Results for the synthetic SPBNs with two auxiliary source domains. 0\% and 10\% of arc modification.}
    \label{fig:010p_modified_synthetic}
\end{figure}

\begin{figure}[!htb]
    \centering
    \subfigure{
        \includegraphics[width=.98\linewidth]{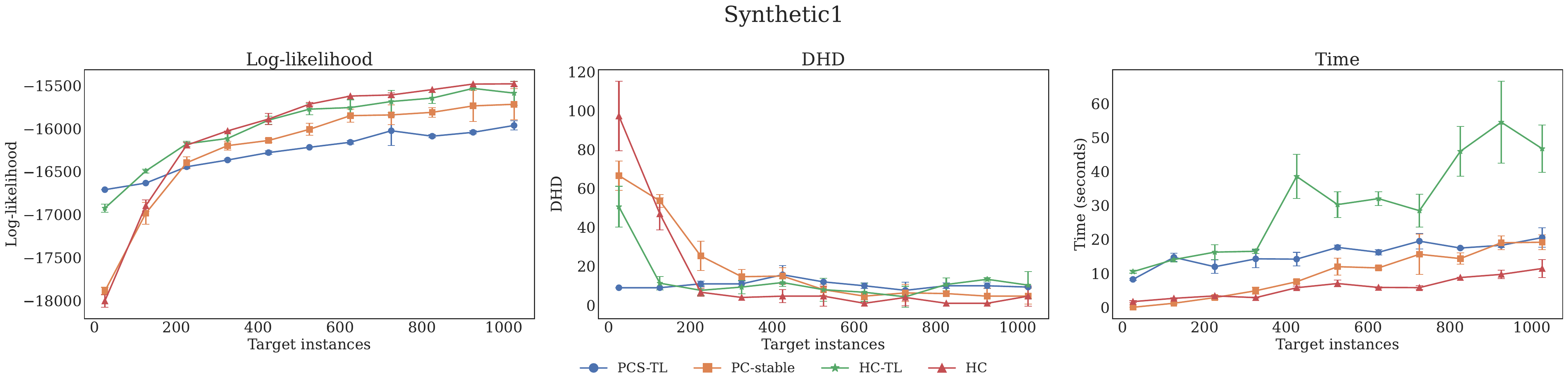}
        \label{fig:s1_51020p}
    }
    \subfigure{
        \includegraphics[width=.98\linewidth]{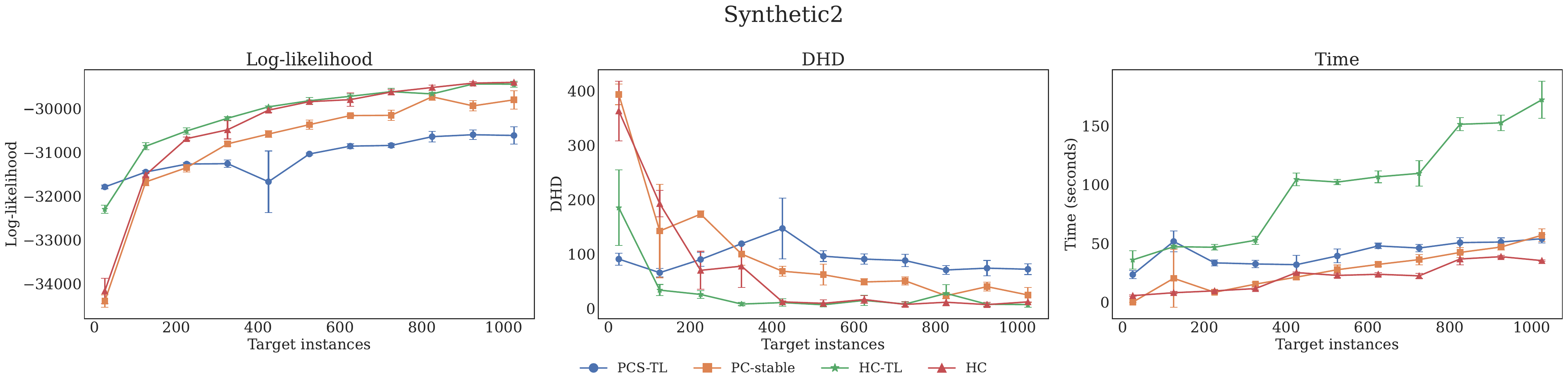}
        \label{fig:s2_51020p}
    }
    \subfigure{
        \includegraphics[width=.98\linewidth]{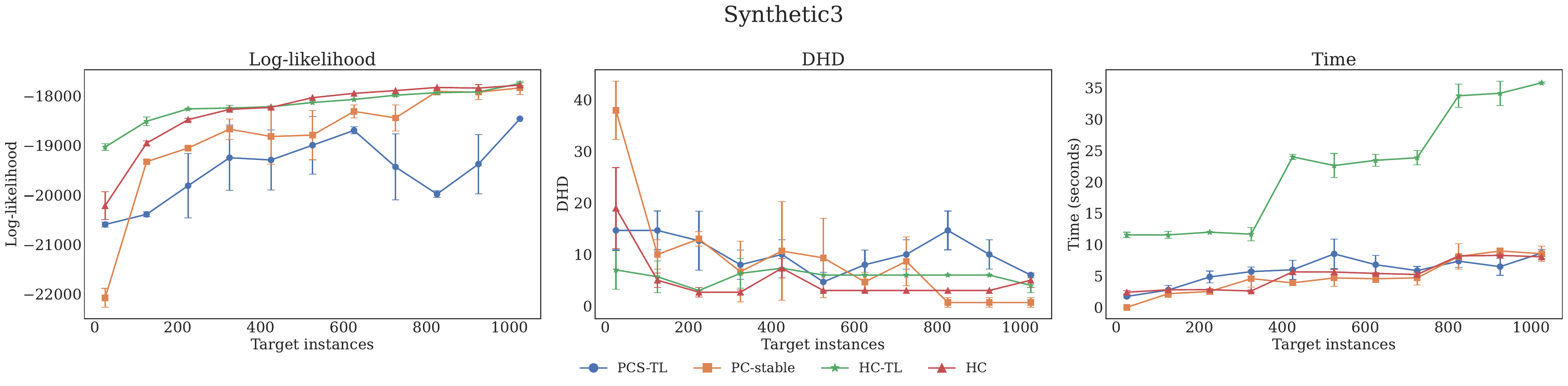}
        \label{fig:s3_51020p}
    }    
    \subfigure{
        \includegraphics[width=.98\linewidth]{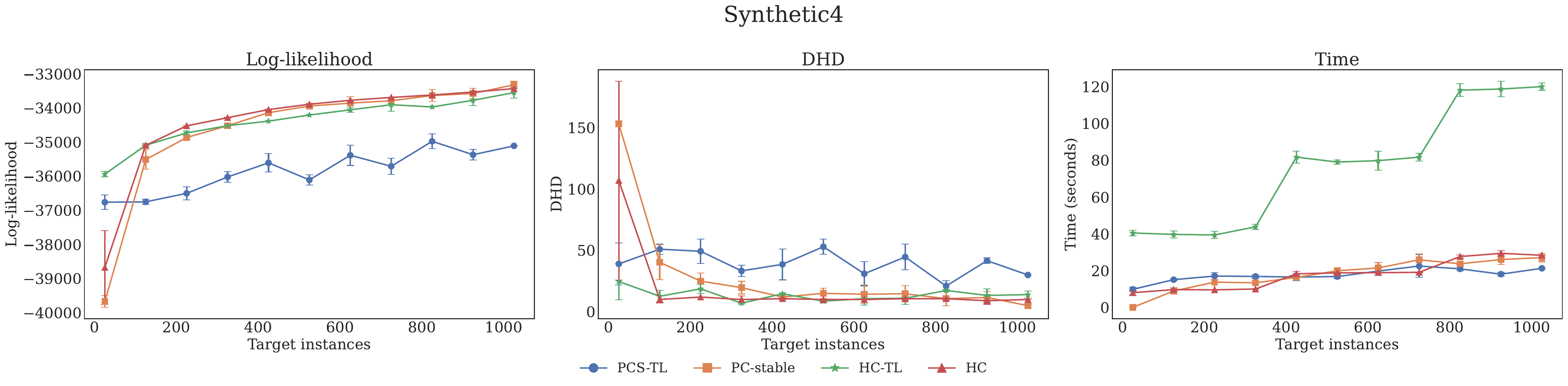}
        \label{fig:s4_51020p}
    }
    \caption{Results for the synthetic SPBNs with three auxiliary source domains. 5\%,  10\% and 20\% of arc modification.}
    \label{fig:51020_modified_synthetic}
\end{figure}

\begin{figure}[!htb]
    \centering
    \subfigure{
        \includegraphics[width=.98\linewidth]{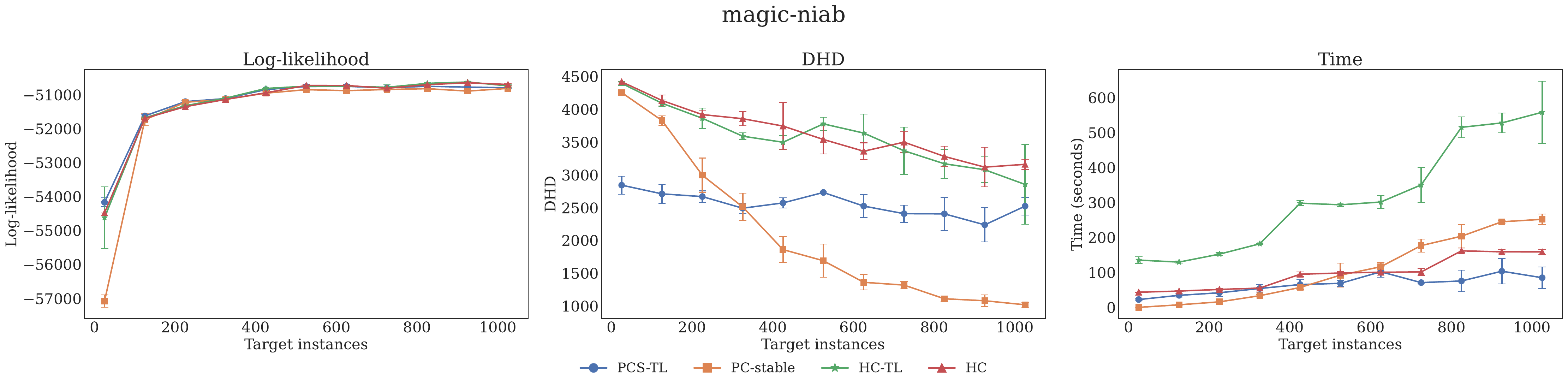}
        \label{fig:magicniab_010p}
    }    
    \subfigure{
        \includegraphics[width=.98\linewidth]{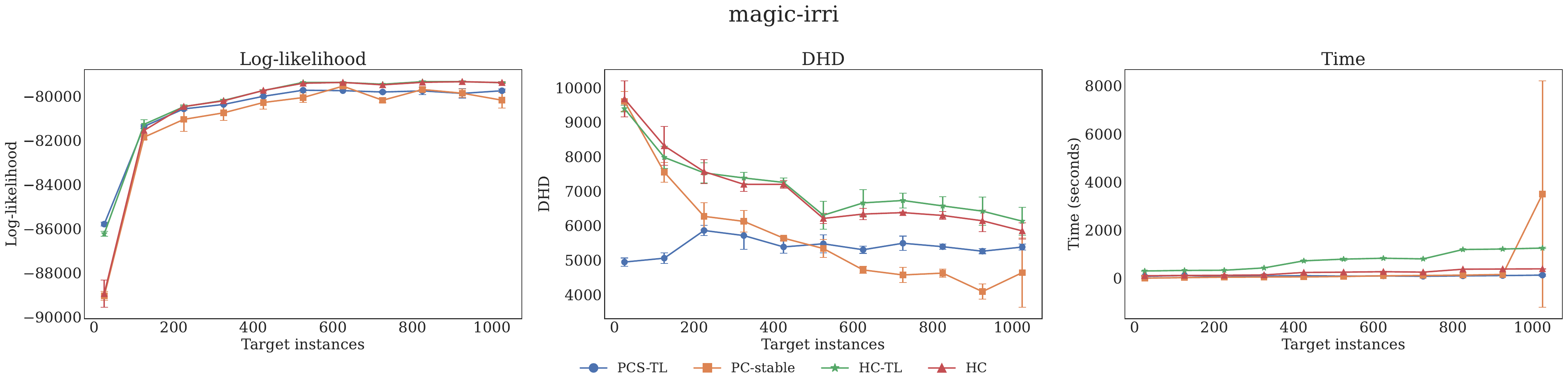}
        \label{fig:magicirri_010p}
    }
    \caption{Results for the bnlearn networks with two auxiliary source domains. 0\% and 10\% of arc modification.}
    \label{fig:010p_modified_bnlearn}
\end{figure}

\begin{figure}[!htb]
    \centering
    \subfigure{
        \includegraphics[width=.98\linewidth]{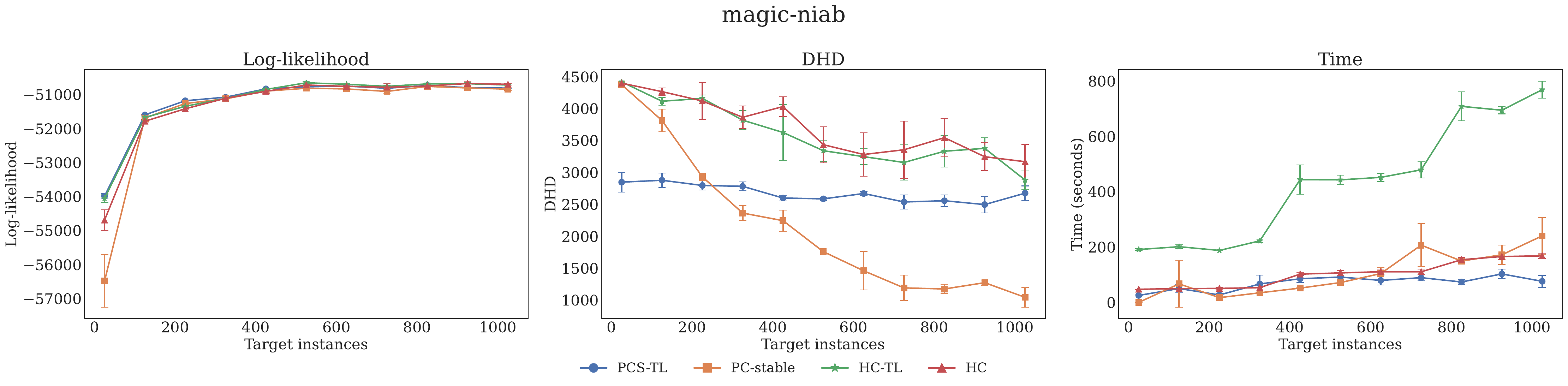}
        \label{fig:magicniab_51020p}
    }    
    \subfigure{
        \includegraphics[width=.98\linewidth]{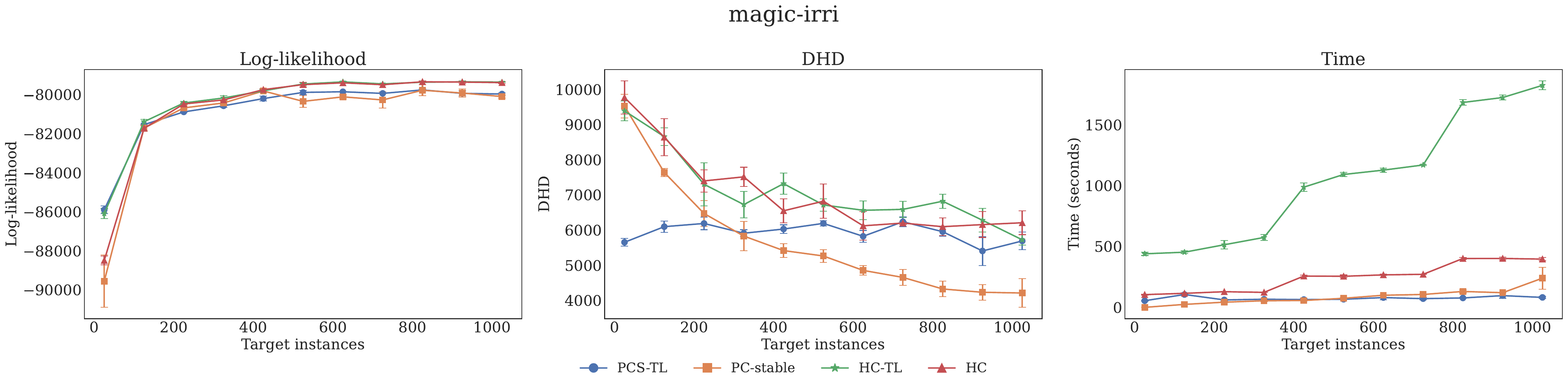}
        \label{fig:magicirri_51020p}
    }
    \caption{Results for the bnlearn networks with three auxiliary source domains. 5\%, 10\% and 20\% of arc modification.}
    \label{fig:51020p_modified_bnlearn}
\end{figure}

In contrast, the target dataset will be sampled from the true Bayesian network directly. Then, we will sample 3000 instances from each source task and incrementally evaluate the models obtained through the transfer learning methodologies and the target alone. Starting at 25 target instances, the increments will be of 100 until 1025 instances. Our intention with this setup is to evaluate the performance of PCS-TL and HC-TL as the number of target instances increases. Transfer learning aims to address situations of data scarcity. Therefore, the target alone should converge or surpass the transfer learning model as more target information is collected. 
Figure \ref{fig:010p_modified_synthetic} and Figure \ref{fig:51020_modified_synthetic} illustrate the results of two and three auxiliary sources for the synthetic (small) SPBNs, while Figure \ref{fig:010p_modified_bnlearn} and Figure \ref{fig:51020p_modified_bnlearn} illustrate the results for the synthetic (medium and large) GBNs.

In all the scenarios, the transfer learning models (shown in blue and green) achieve better results than their non-transfer learning counterparts (shown in orange and red) for both log-likelihood and DHD metrics. Note that when we refer to these metrics, we compare PCS-TL with PC-stable and HC-TL with HC. This is the case for all the experiments where the target task has 25 instances and implies that both algorithms, PCS-TL and HC-TL, achieved closer structures to the ground truth than the models alone. This is achieved despite the data scarcity of the target and the modification of arcs and Gaussian noise of the auxiliary sources. After that, they converge at different rates, which means that both models in both algorithms will eventually reach the same approximate structures, depending on the data complexity (number of variables and data distribution). 

Specifically, for synthetic SPBNs 1 and 2 in Figure \ref{fig:010p_modified_synthetic}, the results of the log-likelihood and DHD of the transfer learning models converge with the target alone around 400 target instances. Similarly, in Figure \ref{fig:51020_modified_synthetic}, the convergence occurs around the 300 instances for SPBNs 1 and 2. However, for synthetic SPBNs 3 and 4, the convergence occurs almost immediately in both Figures. Here, most of the transfer learning models have converged at 125 instances. After that, the log-likelihood and DHD results show that in most cases, the target alone even surpasses the transfer learning counterpart. This surpass is especially evident for the log-likelihood of the PCS-TL models. In contrast, the risk metric (Equation \eqref{eq:risk}) of the HC-TL strategy ensured that the target task alone converges with the transfer learning task without surpassing its network's log-likelihood.

In Figures \ref{fig:010p_modified_bnlearn} and \ref{fig:51020p_modified_bnlearn}, the convergence according to the DHD of the medium and large synthetic GBNs, magic-niab and magic-irri, occurred immediately for HC-TL, and around 300 target instances for PCS-TL. In HC-TL, we can see slightly smaller DHD values than HC at the beginning, and almost identical results for the rest of the time. In contrast, PCS-TL shows clearly better results than its counterpart until the convergence. Then, PC outperforms PCS-TL. In PCS-TL, we appreciate the same behavior as before, but for HC-TL this is roughly half of the instances we saw for synthetic SPBNs 1 and 2. Here, the difference in convergence rates is due to the characteristics of the different networks. All the synthetic SPBNs contain nonparametric distributions (gray nodes) in more than half of their variables. However, the synthetic SPBNs 1 and 2 have up to 3 and 5 parent nodes,  while the synthetic SPBNs 3 and 4 contain only one parent per node, making the synthetic SPBNs 3 and 4 simpler than SPBNs 1 and 2. For magic-niab and magic-irri, we have a much more complex structure than any of the synthetic SPBNs despite having Gaussian nodes. Thus, it becomes much more difficult for both HC and HC-TL to find suitable conditional relationships under data scarcity, considering that they start from an empty graph. Also, the risk metric could be stopping certain arcs from appearing in the transfer learning model to not harm the log-likelihood of the target.

Concerning the execution times, it is clear that the slowest algorithm is HC-TL, followed by PCS-TL. Despite this, the PCS-TL algorithm works in pairs with PC-stable and HC in most cases, as shown in Figures \ref{fig:51020_modified_synthetic}, \ref{fig:010p_modified_bnlearn}, and \ref{fig:51020p_modified_bnlearn}. For HC-TL, the gap with the rest is higher and becomes bigger as the number of target instances increases. The reason is related to the cost of the algorithms. In HC, there are $n(n-1)$ operations in a graph with $n$ variables, whereas each KDE operation has a cost of $O(N^2)$ for a dataset with $N$ instances. Each time $f_{\text{CKDE}}(x_i|\textbf{x}_{\text{Pa}(i)})$ (Equation (\ref{eq:non-parm-cpd})) is computed, it requires two KDE operations, $O(2N^2)$. In HC-TL, this is $O(2N^TN^T)$ for the target, and $O(2N^TsN^s)$ for $s$ sources. Therefore, it requires $O(2N^T(N^T+sN^s))$.

Alternatively, PC-stable performs $n(n-1)/2$ CI tests during the first iteration and an indeterminate number of tests for the rest, since each iteration depends on the previously found conditional independencies. The complexity of the RCoT CI test is $O(N n(\textbf{Z}))$, where $n(\textbf{Z})$ denotes the number of elements in the conditioning set $\textbf{Z}$. In PCS-TL, this is $O(N^T n(\textbf{Z}))$ for the target, and $O(sN^s n(\textbf{Z}))$ for $s$ sources. Therefore, $P_{\text{TL}}(X,Y|\textbf{Z})$ (Equation (\ref{eq:pval_transfer})) requires $O((N^T + sN^s) n(\textbf{Z}))$. Assuming that the number of CI tests does not exploit for a particular setting, $O(2N^T(N^T+sN^s))$ for HC-TL is greater than $O((N^T + sN^s) n(\textbf{Z}))$ for PCS-TL. Thus, increasing the complexity of the dataset will have a bigger impact on HC-TL than the others.

\subsection{UCI Machine Learning repository}
\label{sec:experiment_uci}
For the UCI Machine Learning repository, we selected five unlabeled datasets with continuous variables from different domains. Table \ref{tab:real_datasets} presents the datasets along with their characteristics after the removal of timestamps, discrete columns, and null values. Since the underlying structure of the network for the UCI datasets is not available, we learned a model with 10000 target instances for both HC and PC-stable algorithms. The resulting structure from these two learning processes can be used as a reference to evaluate the DHD on HC-TL and PCS-TL, respectively. HC and PC-stable are two very different approaches, and we consider that each result should be compared with an optimal structure found with the same learning method. To this end, the same experimental setup as in Section \ref{sec:experiment_synthetic} can be followed. The target is learned using small sample sizes to simulate data scarcity, and the resulting structure is compared with a valid reference through the DHD. 

\begin{table}[H]
  \centering
  \begin{tabular}{clrr}
    \toprule
    \textbf{Dataset} & \textbf{Name}  &\textbf{$N$} & \textbf{$n$}  \\
    \midrule
     1 & Single elder home monitoring: gas and position \citep{1_single_elder}                                      & 416153  & 9  \\
     2 & HTRU2 \citep{2_htru2}                                                                                      & 17898   & 8 \\
     3 & Individual household electric power consumption \citep{individual_household_electric_power_consumption_235} & 2049280 & 7  \\
     4 & MAGIC gamma telescope \citep{4_magic_gamma_telescope}                                                      & 19020   & 10 \\
     5 & Appliances energy prediction \citep{5_appliance_energy_predicion}                                          & 19735   & 24 \\
     \bottomrule
    \end{tabular}
\caption{Datasets from the UCI Machine Learning repository.}\label{tab:real_datasets}
\end{table}

\begin{figure}[!htb]
    \centering
    \subfigure{
        \includegraphics[width=.95\linewidth]{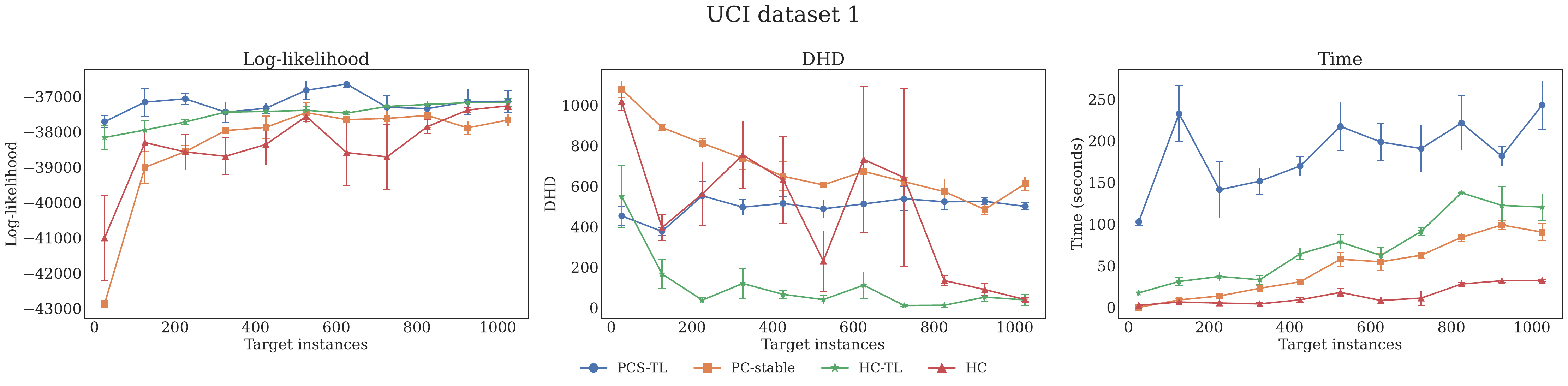}
        \label{fig:1_010p}
    }
    \subfigure{
        \includegraphics[width=.95\linewidth]{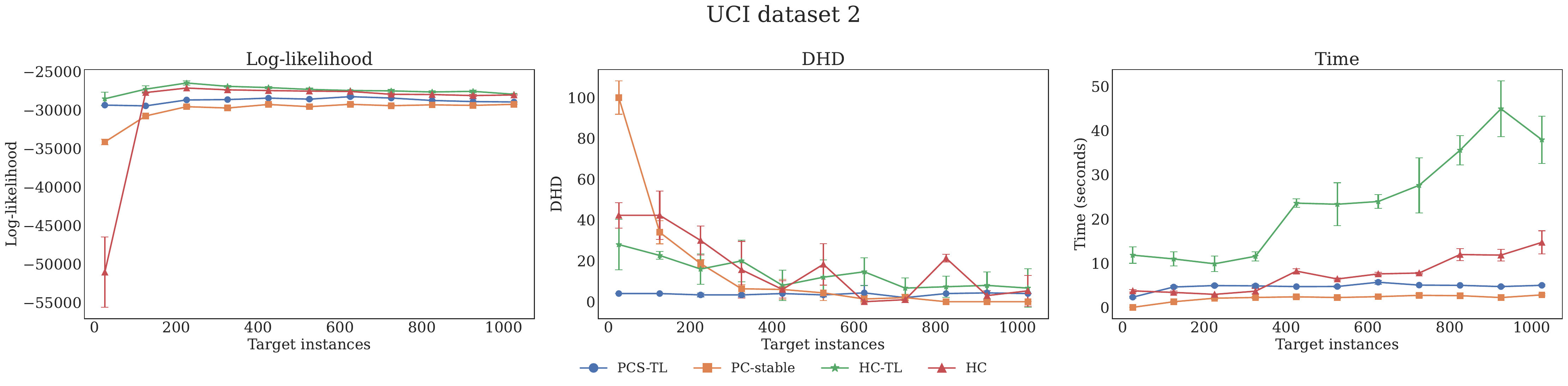}
        \label{fig:2_010p}
    }
    \subfigure{
        \includegraphics[width=.95\linewidth]{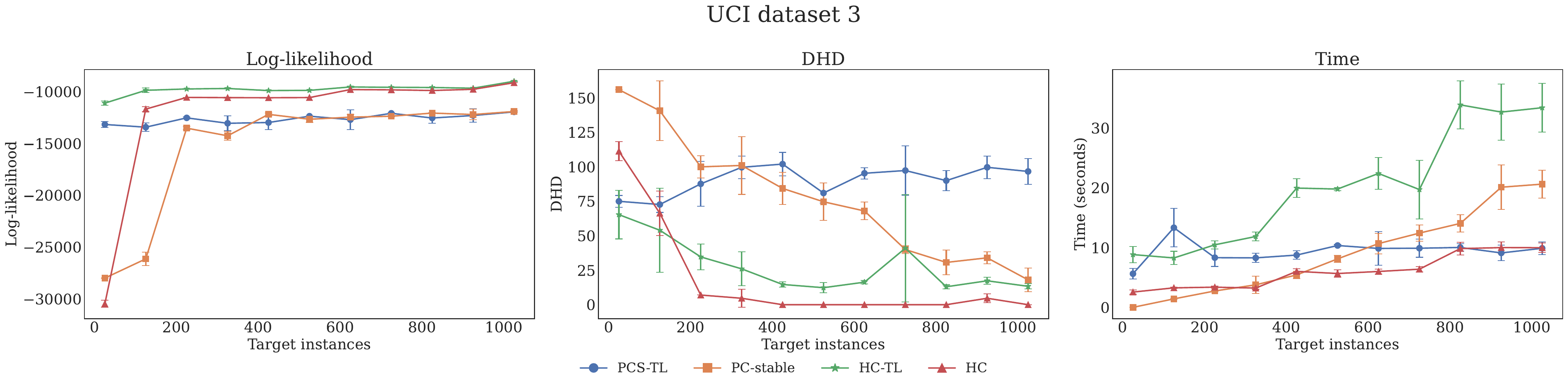}
        \label{fig:3_010p}
    }    
    \subfigure{
        \includegraphics[width=.95\linewidth]{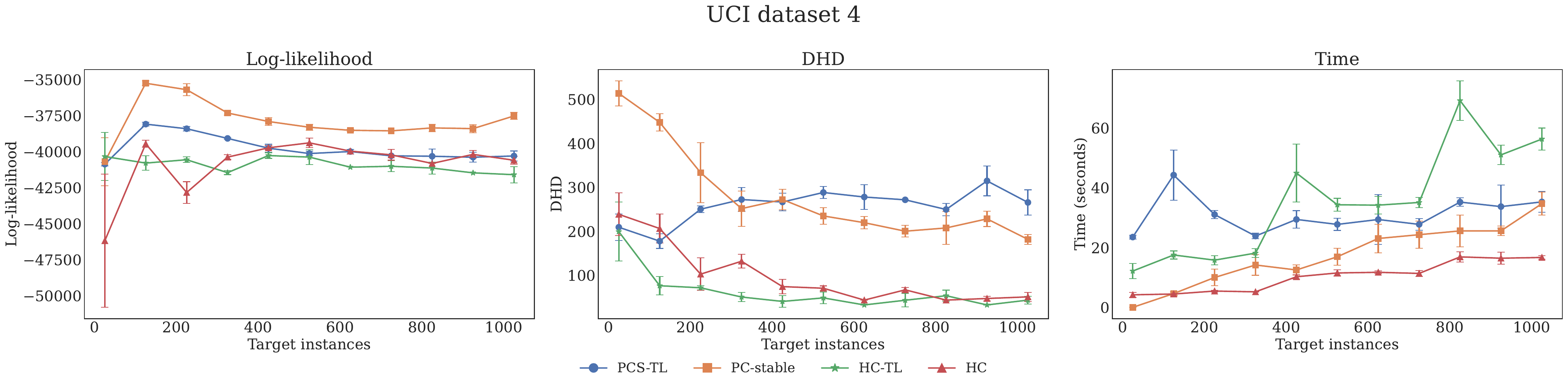}
        \label{fig:4_010p}
    }
    \subfigure{
        \includegraphics[width=.95\linewidth]{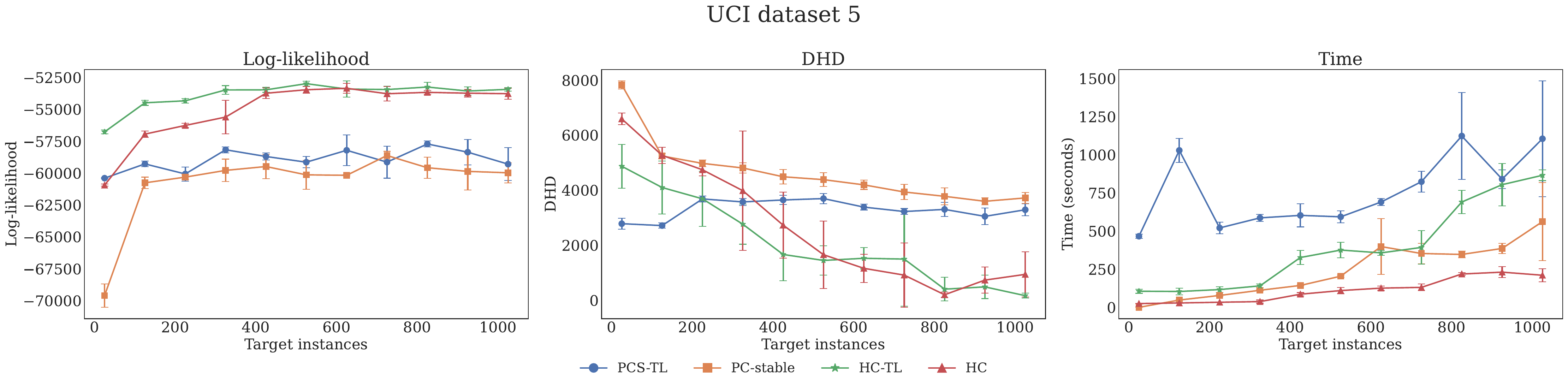}
        \label{fig:5_010p}
    }
    \caption{Results for the UCI datasets with two auxiliary source domains. 0\% and 10\% of arc modification.}
    \label{fig:uci_010p}
\end{figure}

\begin{figure}[!htb]
    \centering
    \subfigure{
        \includegraphics[width=.95\linewidth]{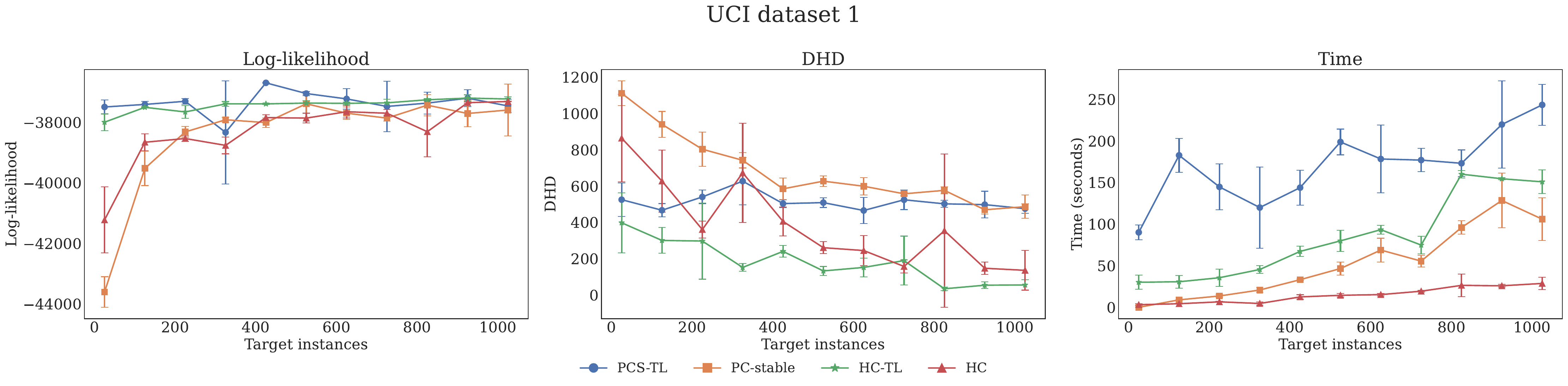}
        \label{fig:1_51020p}
    }
    \subfigure{
        \includegraphics[width=.95\linewidth]{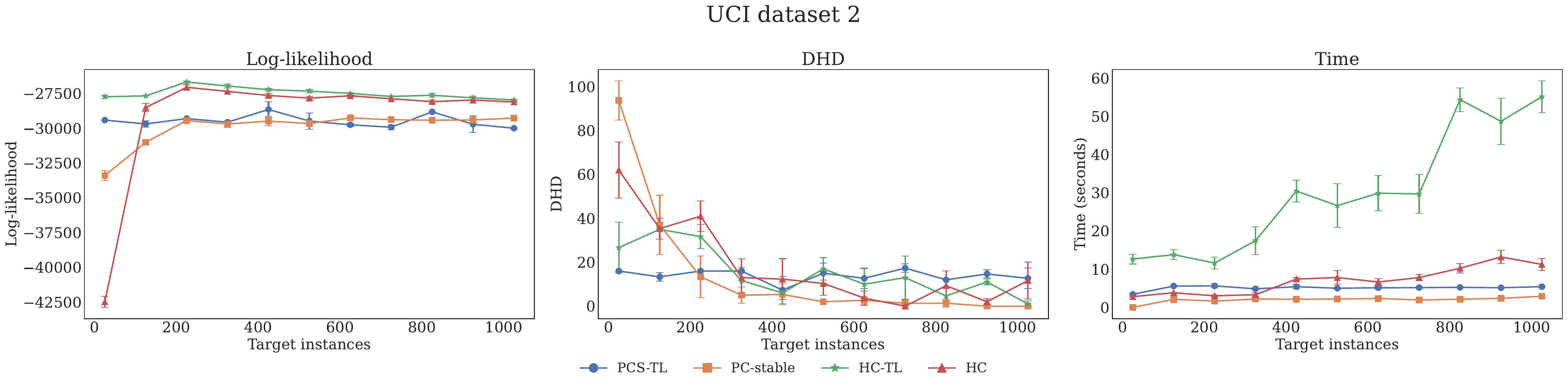}
        \label{fig:2_51020p}
    }
    \subfigure{
        \includegraphics[width=.95\linewidth]{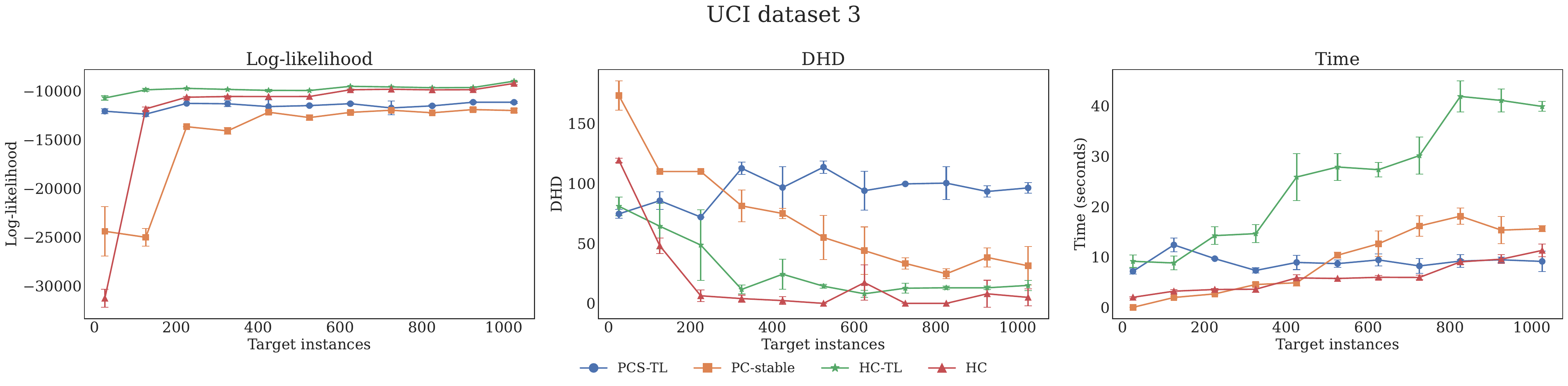}
        \label{fig:3_51020p}
    }    
    \subfigure{
        \includegraphics[width=.95\linewidth]{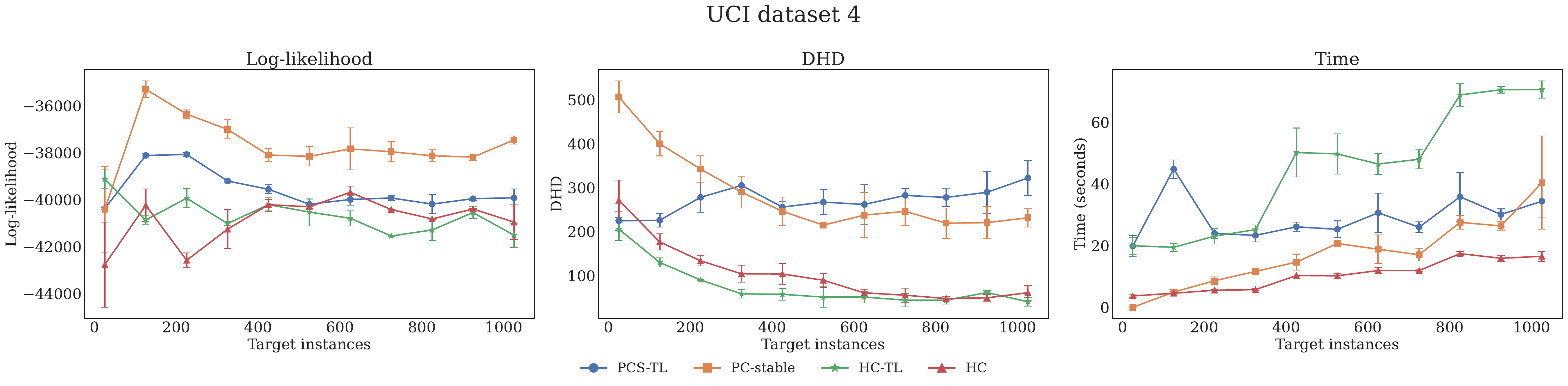}
        \label{fig:4_51020p}
    }
    \subfigure{
        \includegraphics[width=.95\linewidth]{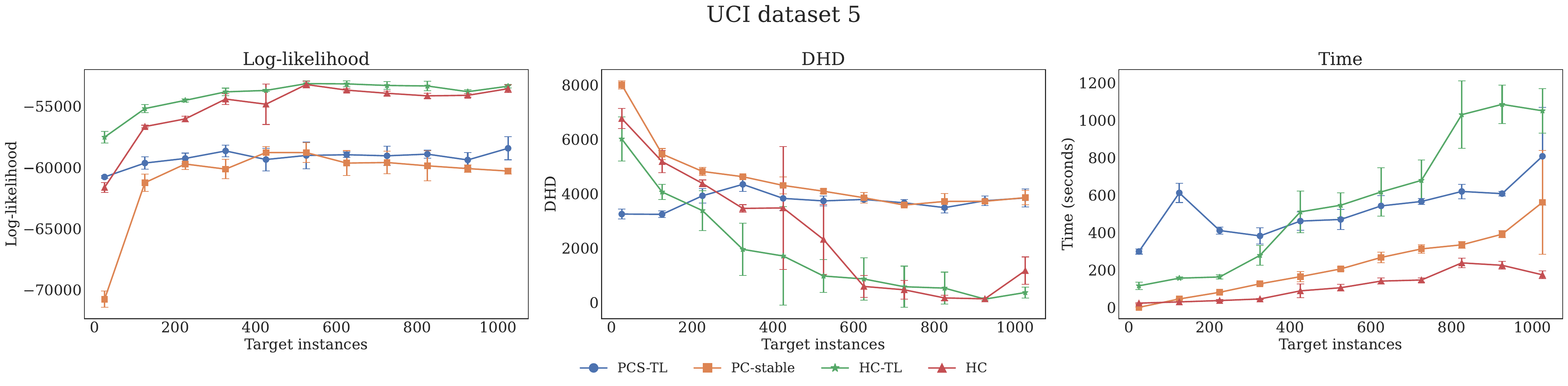}
        \label{fig:5_51020p}
    }
    \caption{Results for the UCI datasets with three auxiliary source domains. 5\%, 10 and 20\% of arc modification.}
    \label{fig:uci_51020p}
\end{figure}

\begin{figure}[!htb]
    \centering
    \textbf{HC} \\
    \subfigure[UCI dataset 1.]{
        \includegraphics[width=0.45\linewidth]{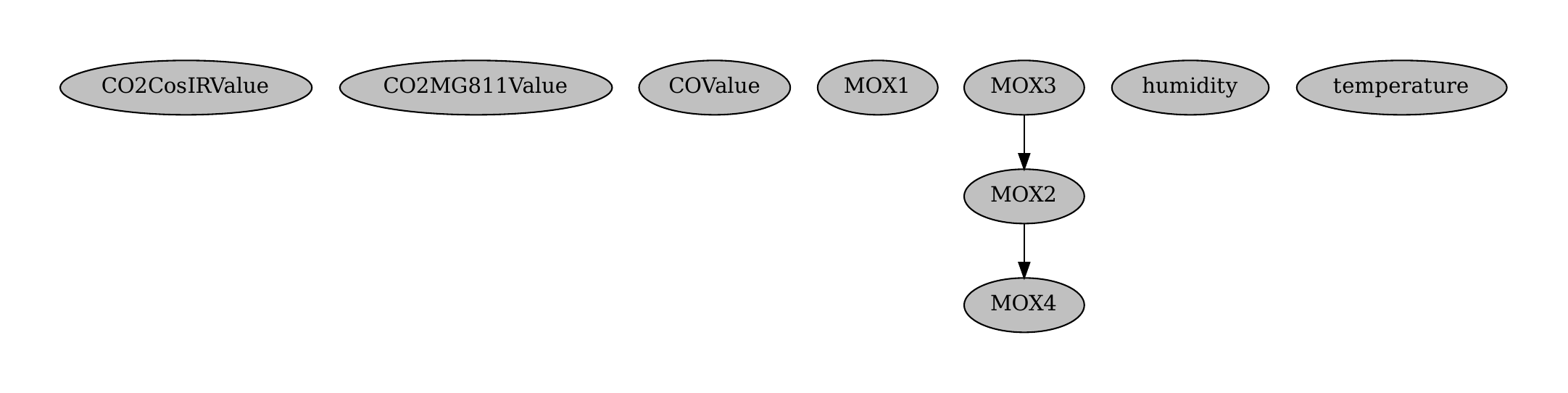}
        \label{fig:hc25-uci1}
    }
    \hfill
    \subfigure[UCI dataset 2.]{
        \includegraphics[width=0.35\linewidth]{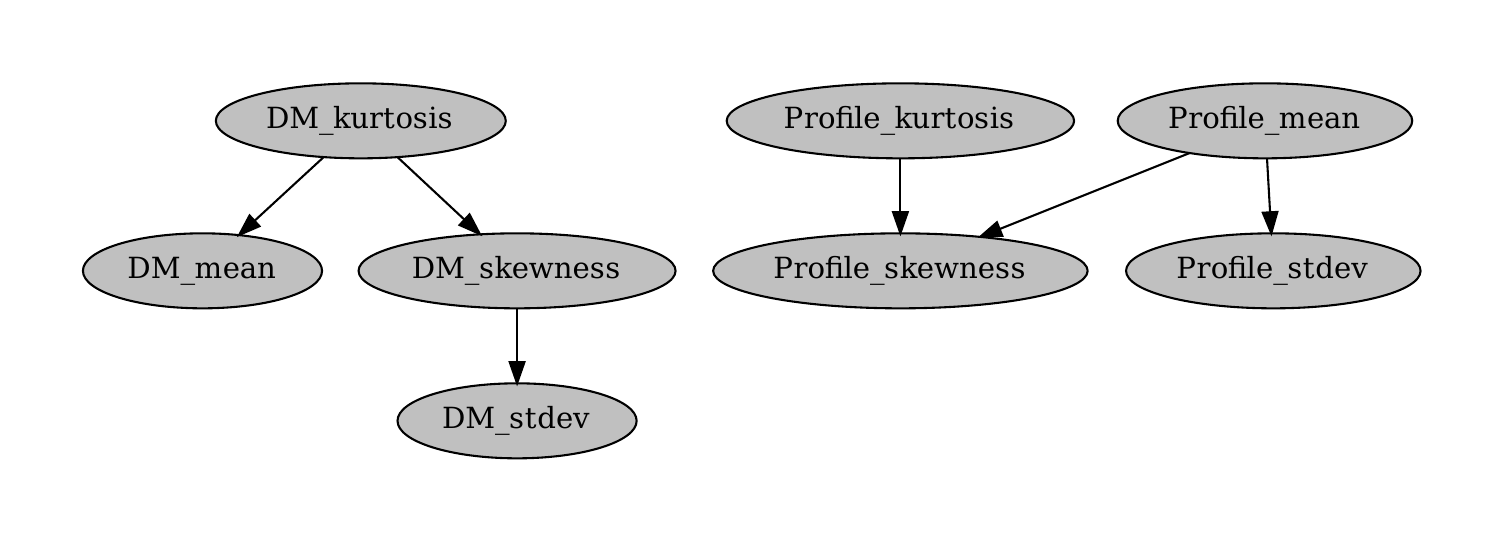}
        \label{fig:hc25-uci2}
    }
    \hfill
    \subfigure[UCI dataset 3.]{
        \includegraphics[width=0.45\linewidth]{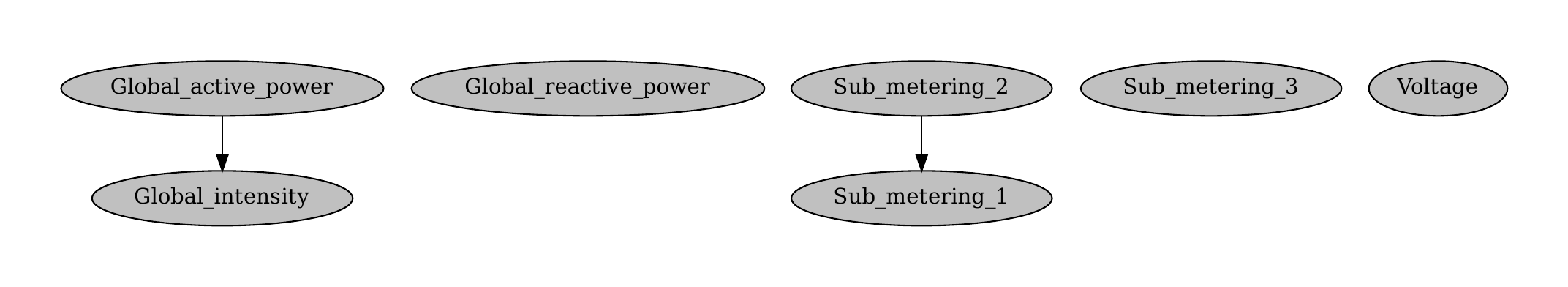}
        \label{fig:hc25-uci3}
    }    
    \hfill
    \subfigure[UCI dataset 4.]{
        \includegraphics[width=0.45\linewidth]{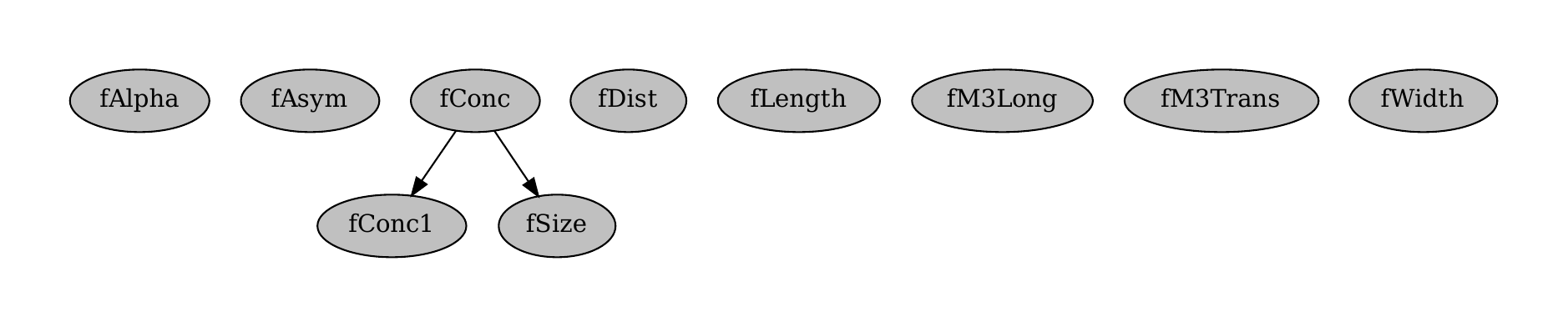}
        \label{fig:hc25-uci4}
    }\hfill
    \subfigure[UCI dataset 5.]{
        \includegraphics[width=0.63\linewidth]{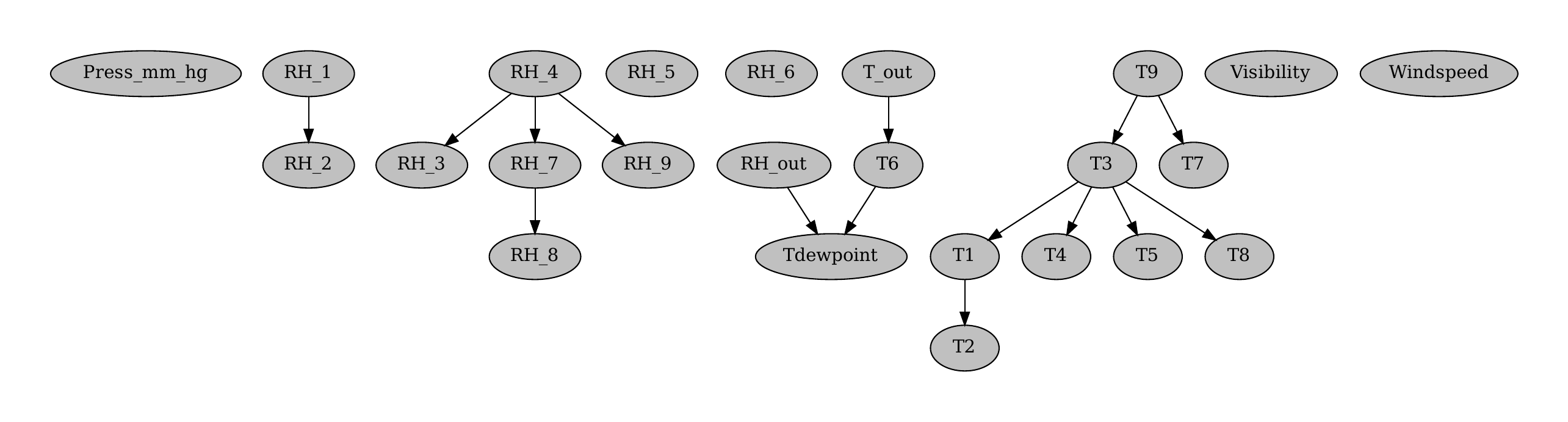}
        \label{fig:hc25-uci5}
    }
    
    \vspace{0.7cm} \textbf{HC-TL} \\
    \hfill
    \subfigure[UCI dataset 1.]{
        \includegraphics[width=0.26\linewidth]{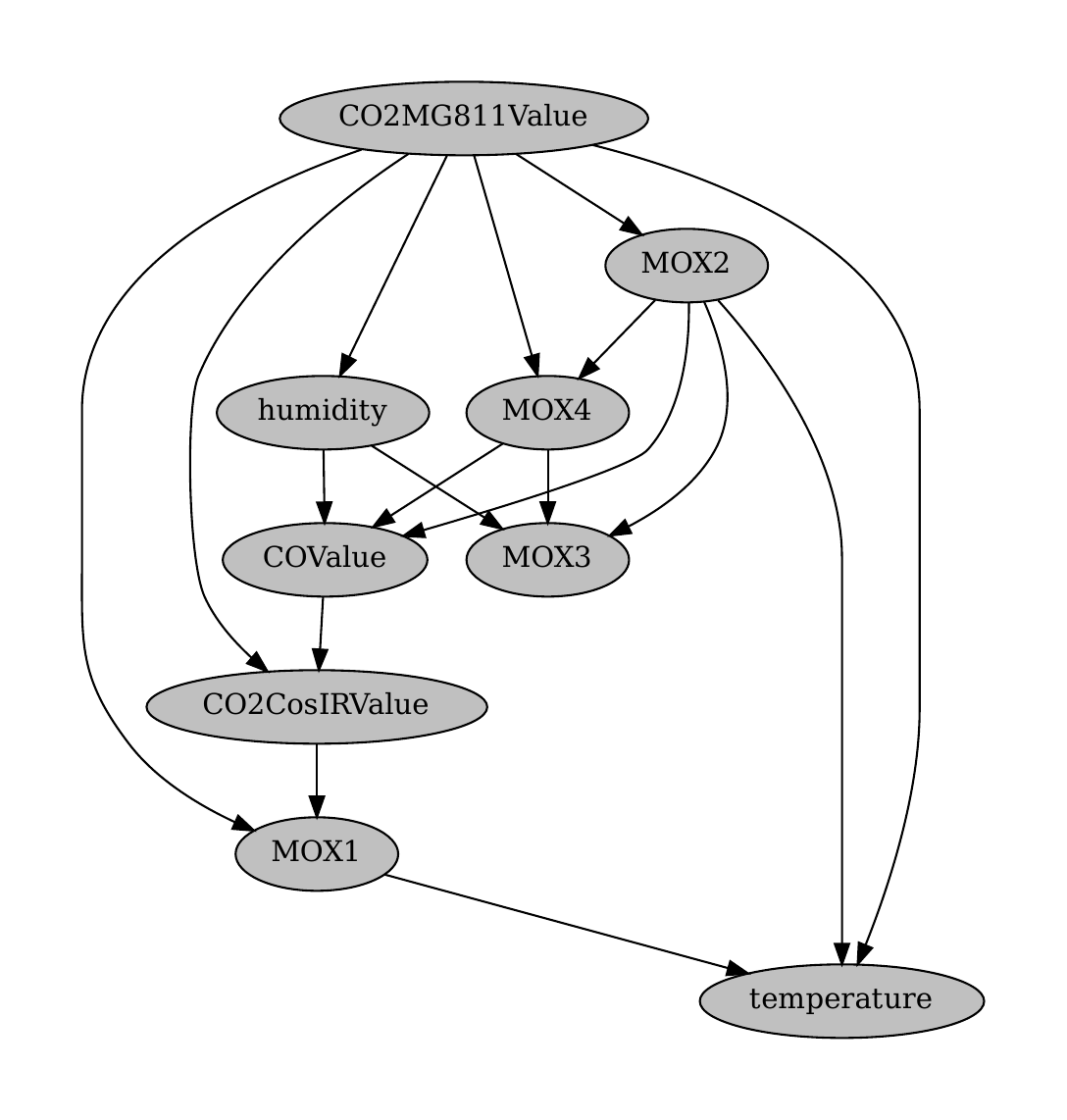}
        \label{fig:hctransfer25-uci1}
    }
    \hfill
    \subfigure[UCI dataset 2.]{
        \includegraphics[width=0.25\linewidth]{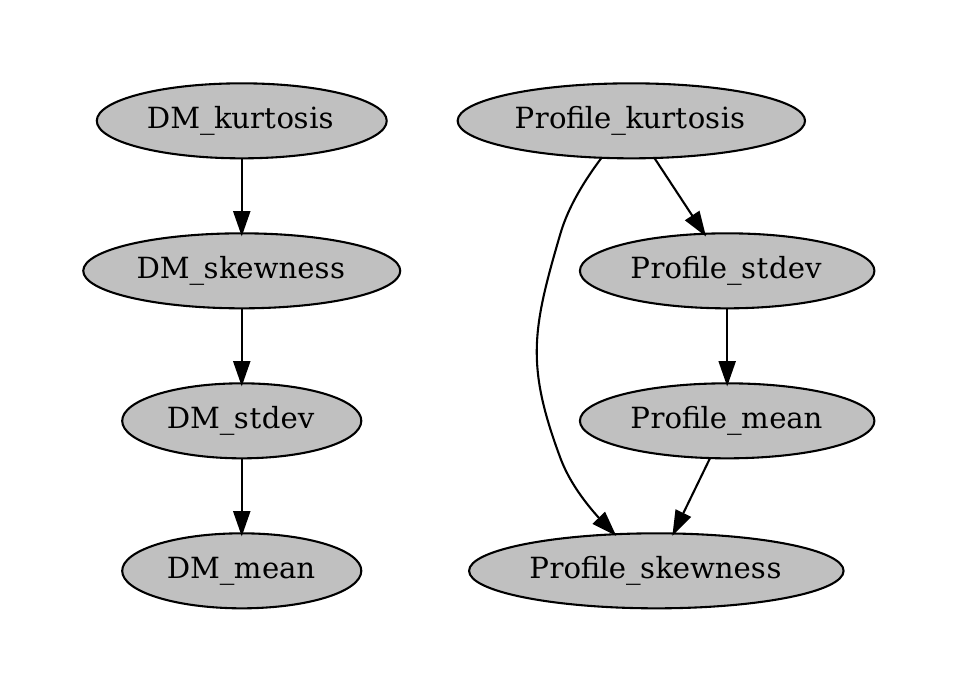}
        \label{fig:hctransfer25-uci2}
    }
    \hfill
    \subfigure[UCI dataset 3.]{
        \includegraphics[width=0.37\linewidth]{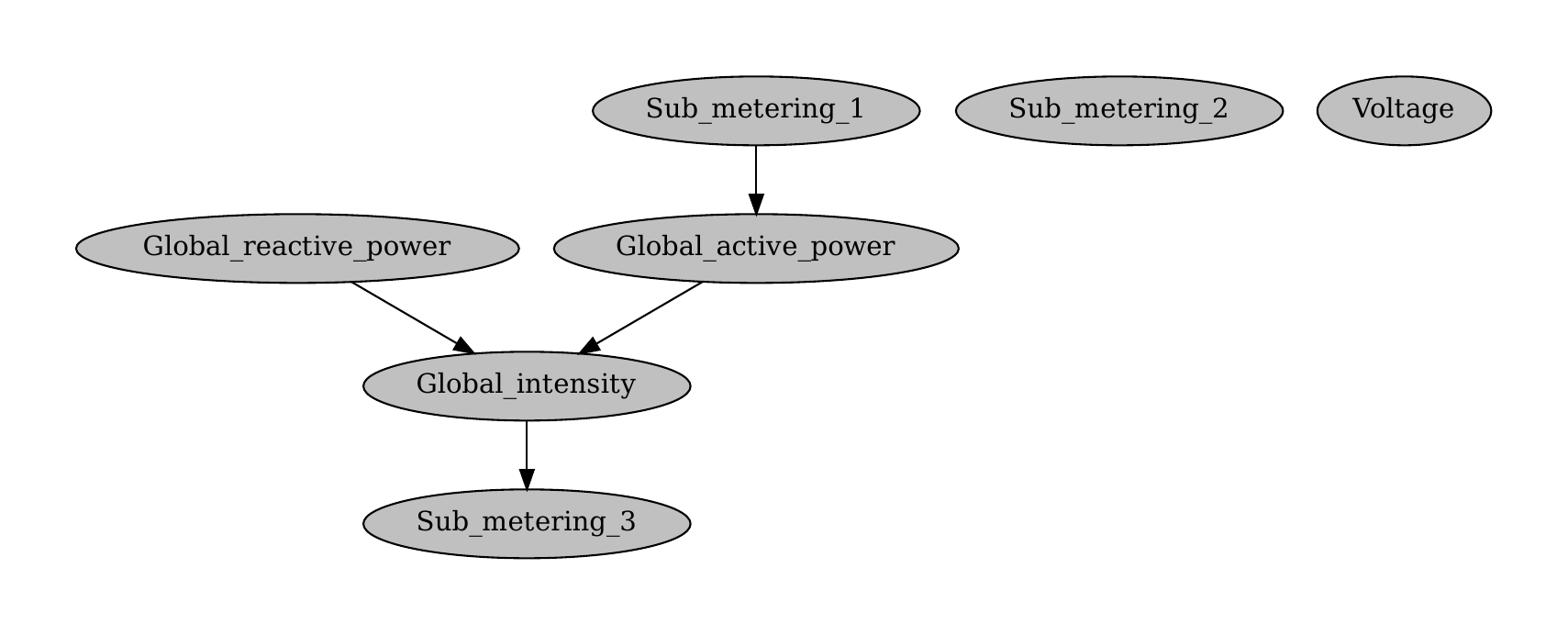}
        \label{fig:hctransfer25-uci3}
    }    
    \hfill
    \subfigure[UCI dataset 4.]{
        \includegraphics[width=0.32\linewidth]{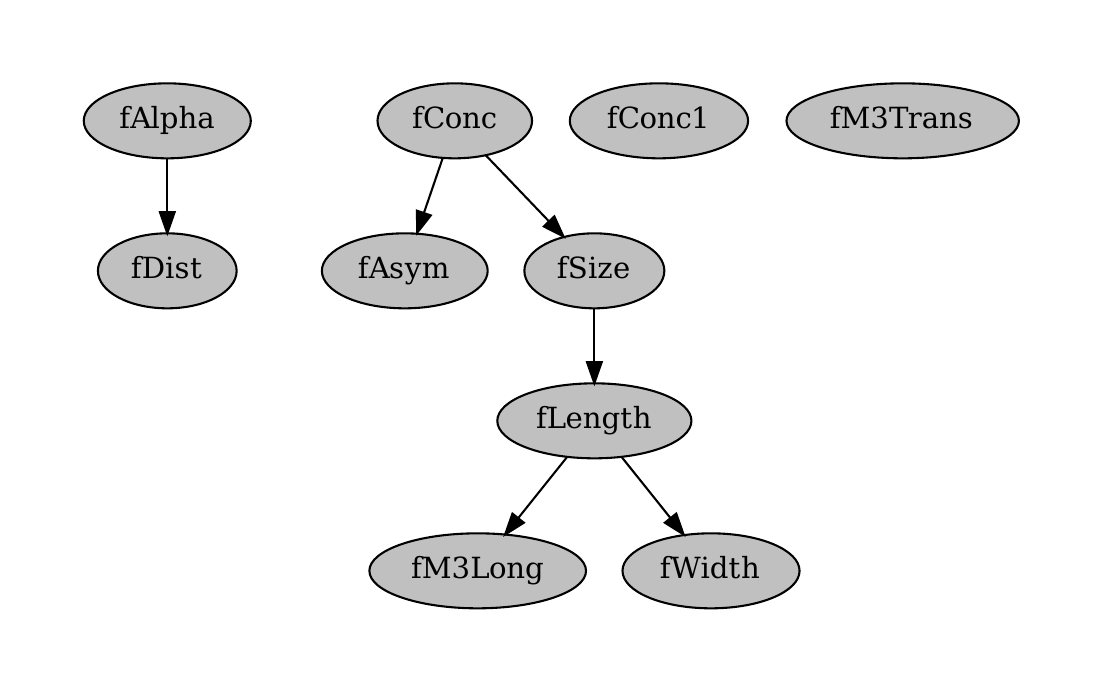}
        \label{fig:hctransfer25-uci4}
    }
    \hfill
     \subfigure[UCI dataset 5.]{
        \includegraphics[width=0.42\linewidth]{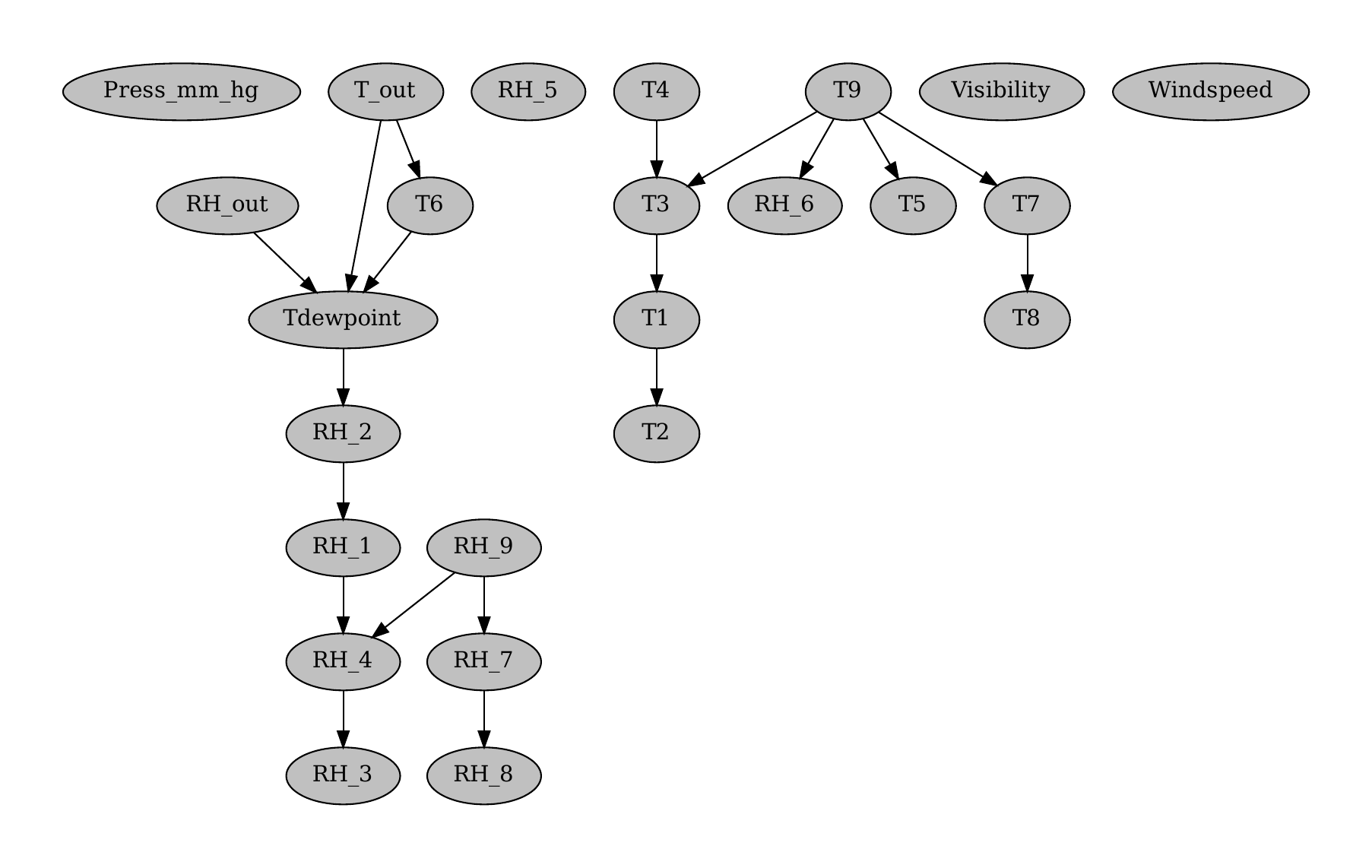}
        \label{fig:hctramsfer25-uci5}
    }
    \caption{Average UCI structures for the HC and HC-TL algorithms with 25 target instances. Results for 3 auxiliary sources with 5\%, 10\% and 20\% of arc modification.}
    \label{fig:hctransfer25-uci}
\end{figure}

\begin{figure}[!htb]
    \centering
    \textbf{PC-stable}\\
    \subfigure[UCI dataset 1.]{
        \includegraphics[width=0.46\linewidth]{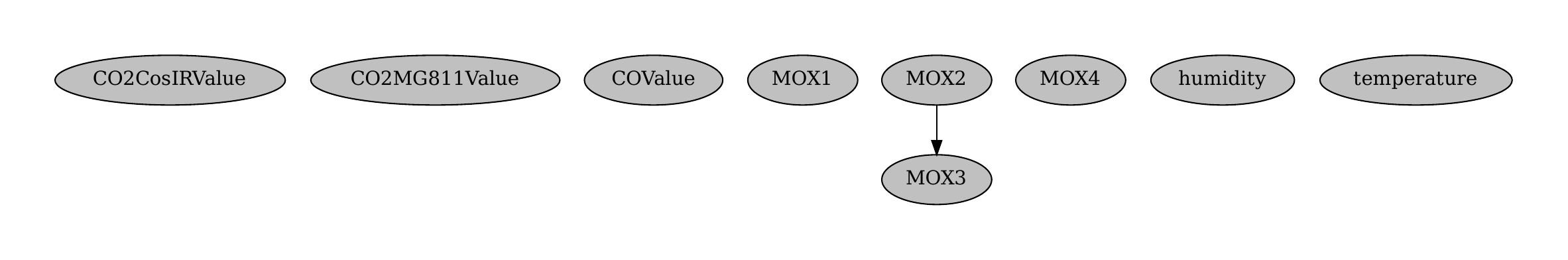}
        \label{fig:pcot25-uci1}
    }
    \hfill
    \subfigure[UCI dataset 2.]{
        \includegraphics[width=0.4\linewidth]{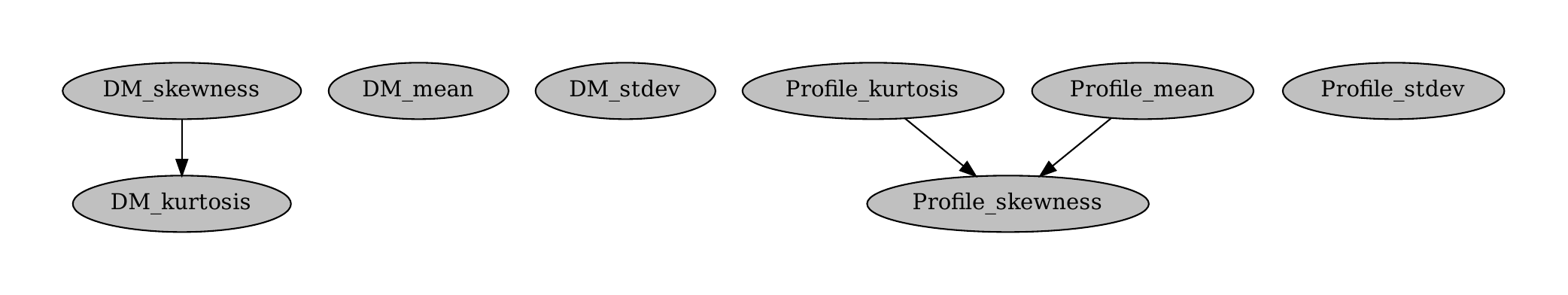}
        \label{fig:pcot25-uci2}
    }
    \hfill
    \subfigure[UCI dataset 3.]{
        \includegraphics[width=0.49\linewidth]{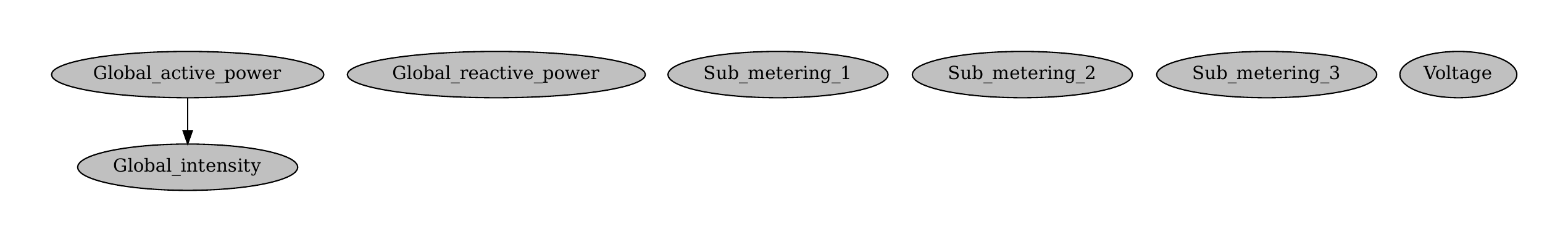}
        \label{fig:pcot25-uci3}
    }    
    \hfill
    \subfigure[UCI dataset 4.]{
        \includegraphics[width=0.45\linewidth]{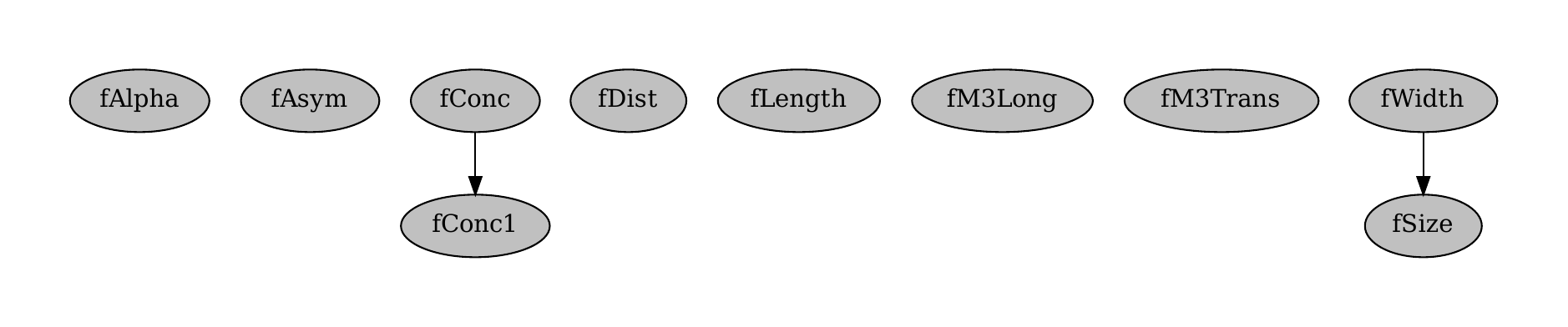}
        \label{fig:pcot25-uci4}
    }
    \hfill
     \subfigure[UCI dataset 5.]{
        \includegraphics[width=0.95\linewidth]{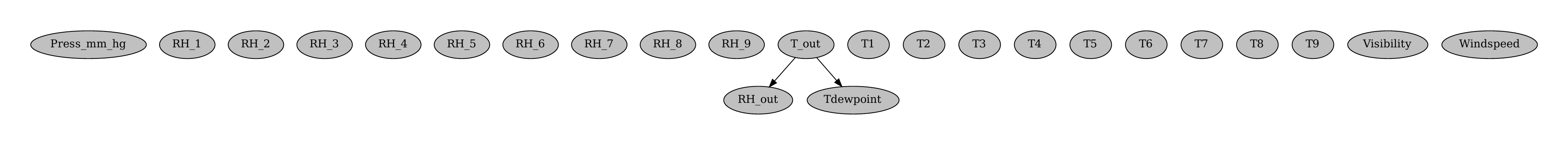}
        \label{fig:pcot25-uci5}
    }
    
    \vspace{0.7cm} \textbf{PCS-TL} \\
    \hfill
    \subfigure[UCI dataset 1.]{
        \includegraphics[width=0.3\linewidth]{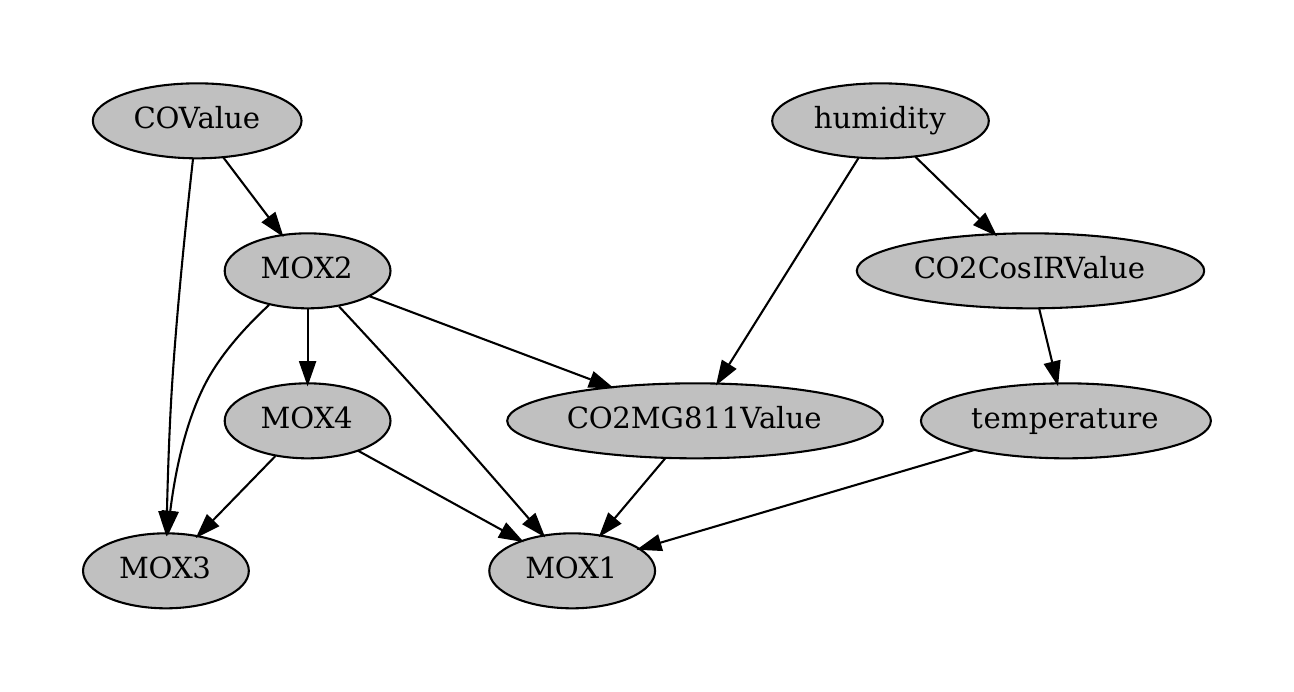}
        \label{fig:pcotransfer25-uci1}
    }
    \hfill
    \subfigure[UCI dataset 2.]{
        \includegraphics[width=0.35\linewidth]{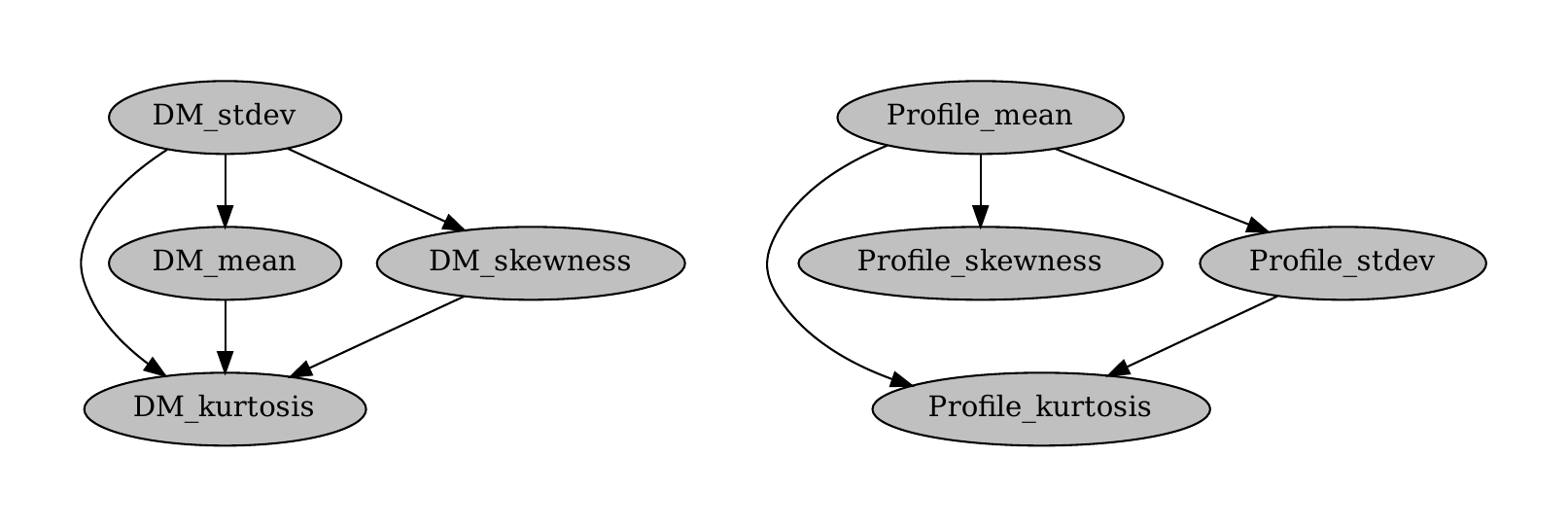}
        \label{fig:pcotransfer25-uci2}
    }
    \hfill
    \subfigure[UCI dataset 3.]{
        \includegraphics[width=0.28\linewidth]{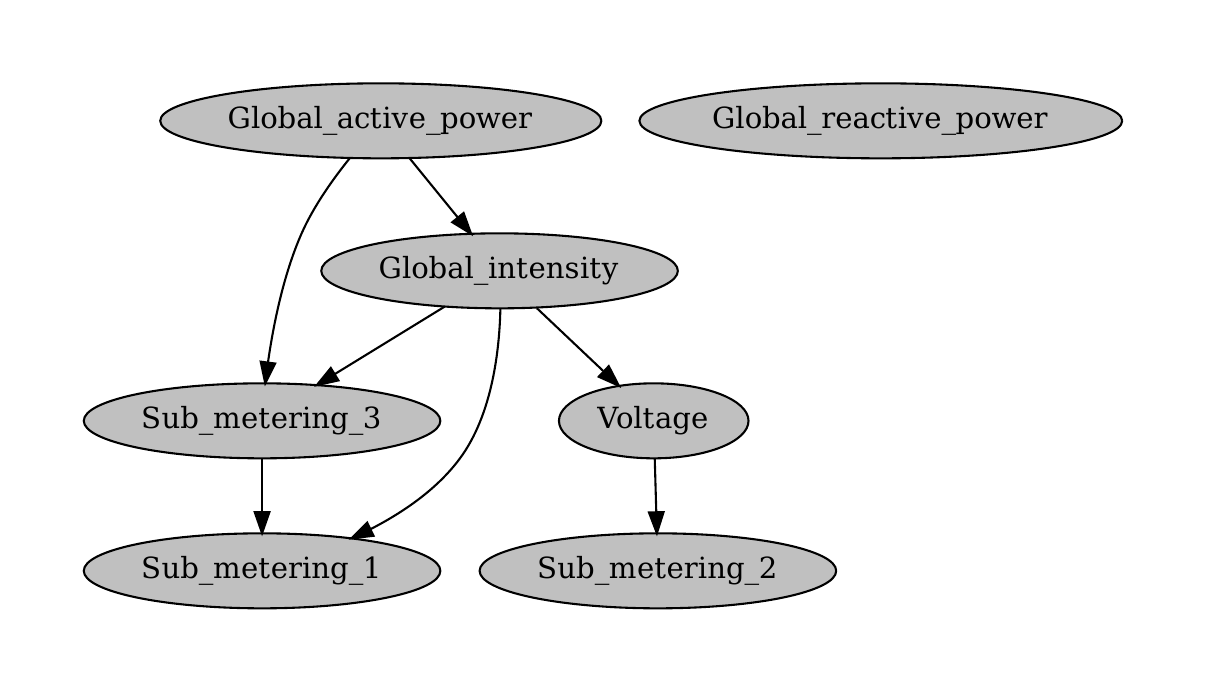}
        \label{fig:pcotransfer25-uci3}
    }    
    \hfill
    \subfigure[UCI dataset 4.]{
        \includegraphics[width=0.3\linewidth]{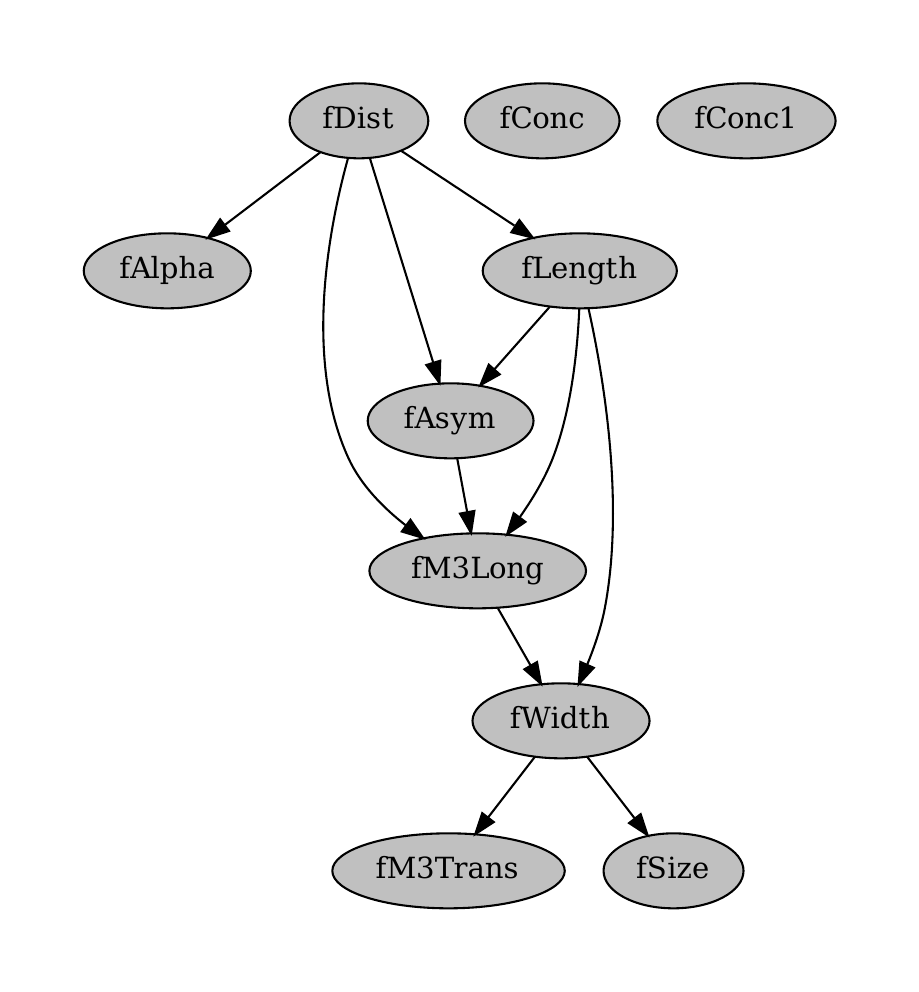}
        \label{fig:pcotransfer25-uci4}
    }
    \hfill
     \subfigure[UCI dataset 5.]{
        \includegraphics[width=0.28\linewidth]{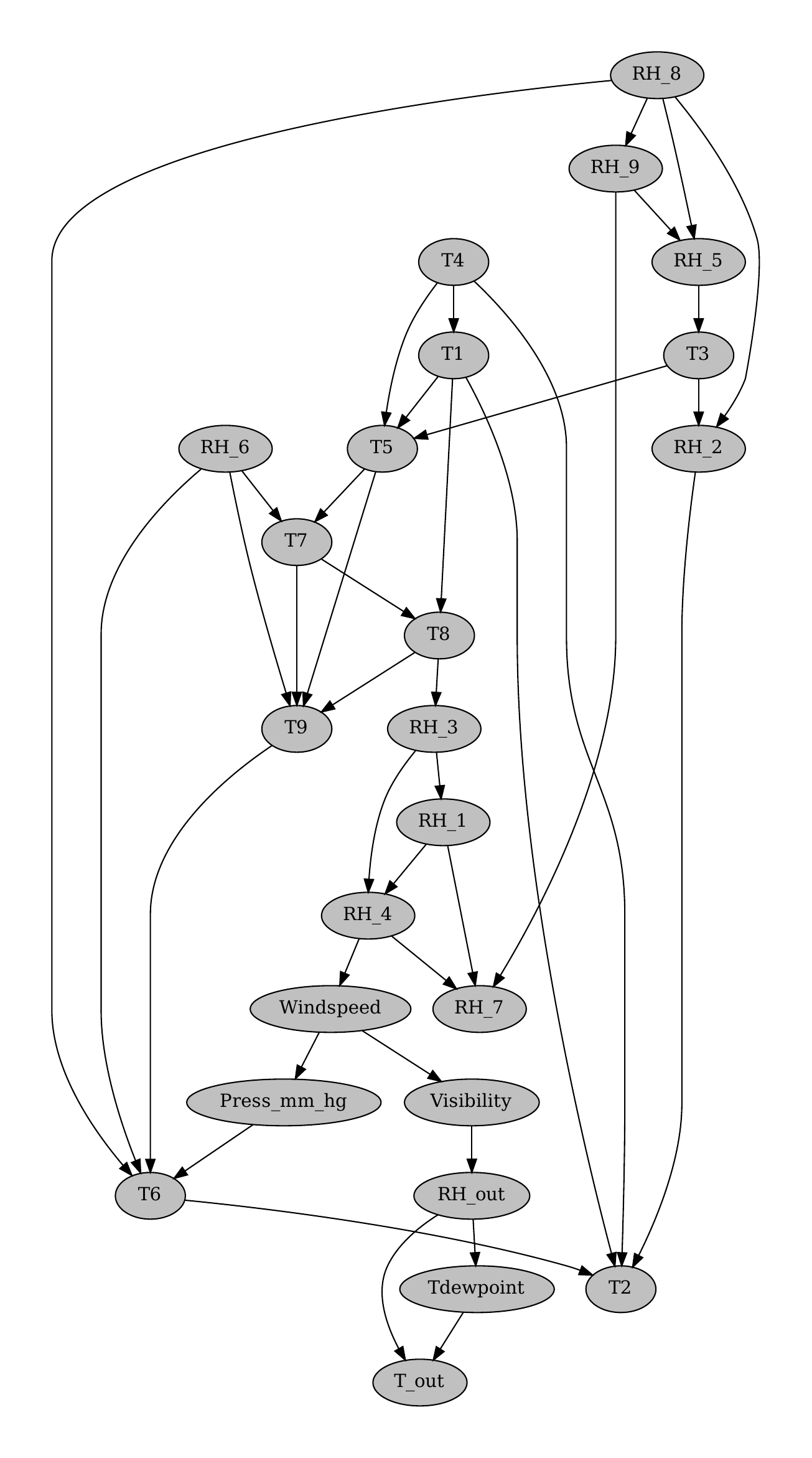}
        \label{fig:pcotransfer25-uci5}
    }
    \caption{Average UCI structures for the PC-stable and PCS-TL algorithms with 25 target instances. Results for 3 auxiliary sources with 5\%, 10\% and 20\% of arc modification.}
    \label{fig:pcotransfer25-uci}
\end{figure}

Figures \ref{fig:uci_010p} and \ref{fig:uci_51020p} show the results for two and three auxiliary sources, respectively, while Figures \ref{fig:hctransfer25-uci} and \ref{fig:pcotransfer25-uci} show the most representative network structures for the different algorithms with 25 target instances and three auxiliary sources. These structures include only the arcs that appeared in at least two of the three runs.
With these images, we can roughly derive the same conclusions as in the previous section. The five transfer learning UCI networks showed equal or better log-likelihood and DHD metrics than the target alone for both algorithms at 25 target instances. However, the difference between the constraint-based models is greater than that between the score-based models, which indicates that PCS-TL produced more complex structures than the HC-TL method for these datasets. Here, the models alone converge with the transfer learning models between 200 and 500 target instances, depending on the data complexity. Thus, the datasets with fewer variables converge earlier than those with a higher number, especially for the score-based networks. There is an exception to this in the log-likelihood of the UCI dataset 4, where both constraint-based and score-based algorithms converge almost immediately despite having more variables than the UCI datasets 2 and 3. 

On the other hand, the execution times of HC-TL and PCS-TL for the UCI datasets also indicate that HC-TL is the slowest algorithm in most cases. There is an exception to this in the results of datasets 1 and 5. Specifically, for dataset 1 in both figures and dataset 5 in Figure \ref{fig:uci_010p}. However, we can argue that the number of CI tests increased the required time in these settings, as PCS-TL was barely affected by the increase in instances. In fact, the times of HC-TL later converge to the same approximate marks, which confirms that it is the most affected by the domain's complexity.

\subsection{Bergmann-Hommel critical difference diagram}
\label{sec:experiment_bergmann}
\begin{figure}[!ht]
    \centering
    \subfigure[Two auxiliary sources with $0\%$ and $10\%$ of the arcs modified]{
        \includegraphics[width=0.9\linewidth]{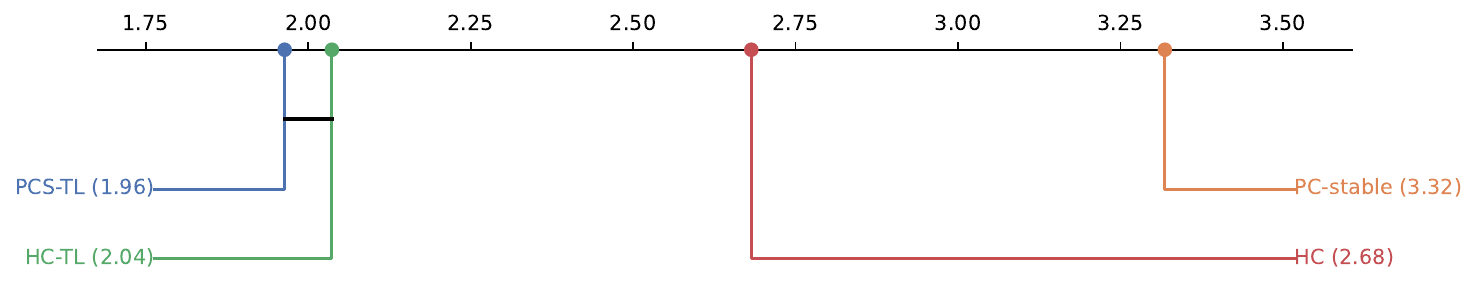}
        \label{fig:DHD_010p_cd}
    }    
    \subfigure[Three auxiliary sources with $5\%$, $10\%$ and $20\%$ of the arcs modified]{
        \includegraphics[width=0.9\linewidth]{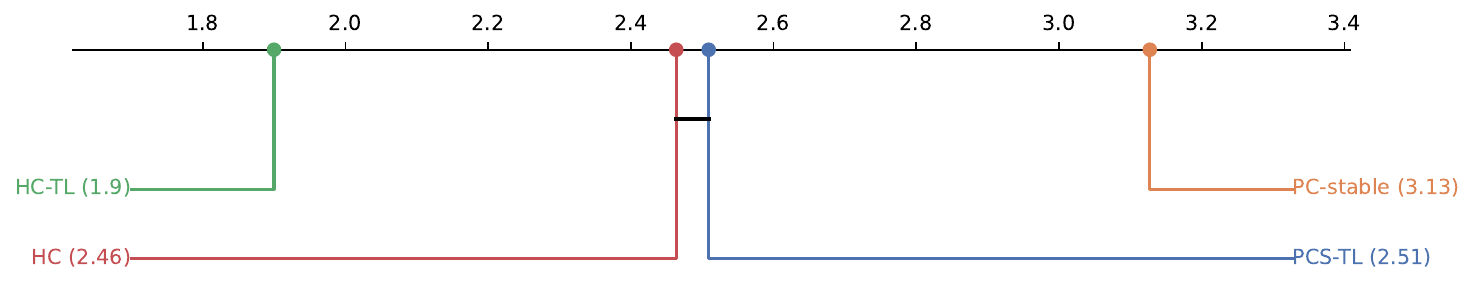}
        \label{fig:DHD_51020p_cd}
    }
    \caption{Critical difference diagram for the network's DHD results. Evaluation for less than $525$ target instances.}
    \label{fig:DHD_cd}
\end{figure}

\begin{figure}[!ht]
    \centering
    \subfigure[Two auxiliary sources with $0\%$ and $10\%$ of the arcs modified]{
        \includegraphics[width=0.44\linewidth]{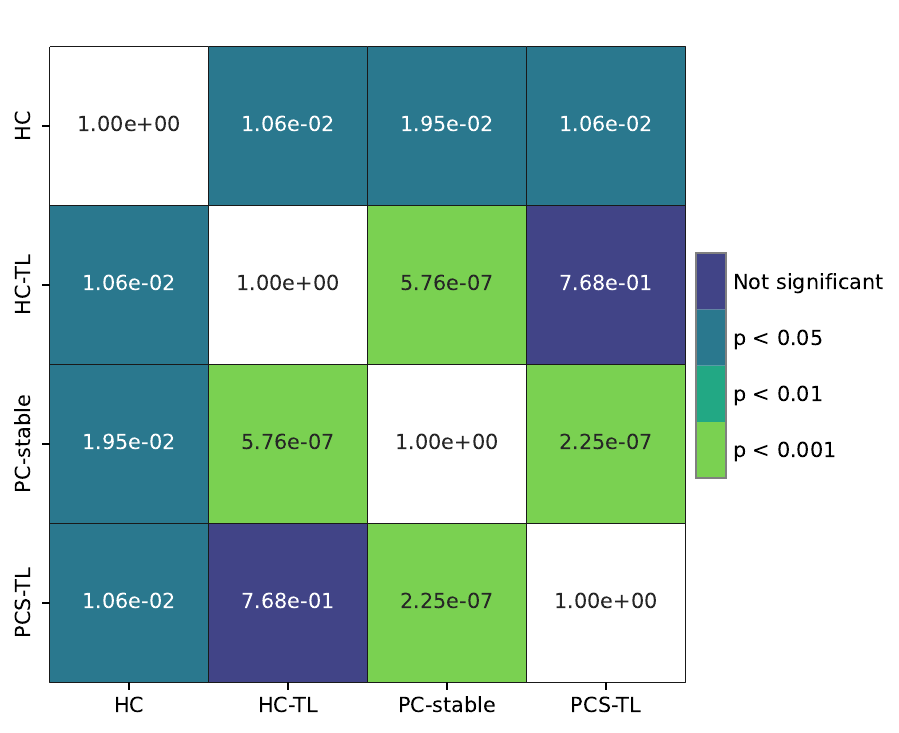}
        \label{fig:DHD_010p_heat}
    }    \hfill
    \subfigure[Three auxiliary sources with $5\%$, $10\%$ and $20\%$ of the arcs modified]{
        \includegraphics[width=0.44\linewidth]{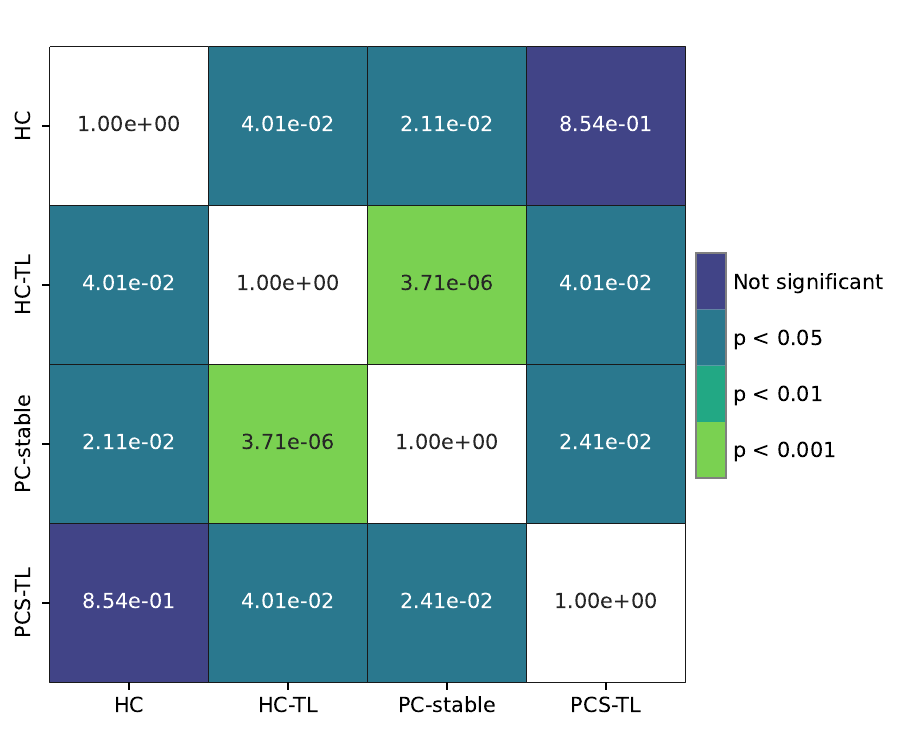}
        \label{fig:DHD_51020p_heat}
    }
    \caption{
    Significance heatmap with $p$-values for the network's DHD results. Evaluation for less than $525$ target instances.}
    
    \label{fig:DHD_heat}
\end{figure}

\begin{figure}[!ht]
    \centering
    \subfigure[Two auxiliary sources with $0\%$ and $10\%$ of the arcs modified]{
        \includegraphics[width=0.9\linewidth]{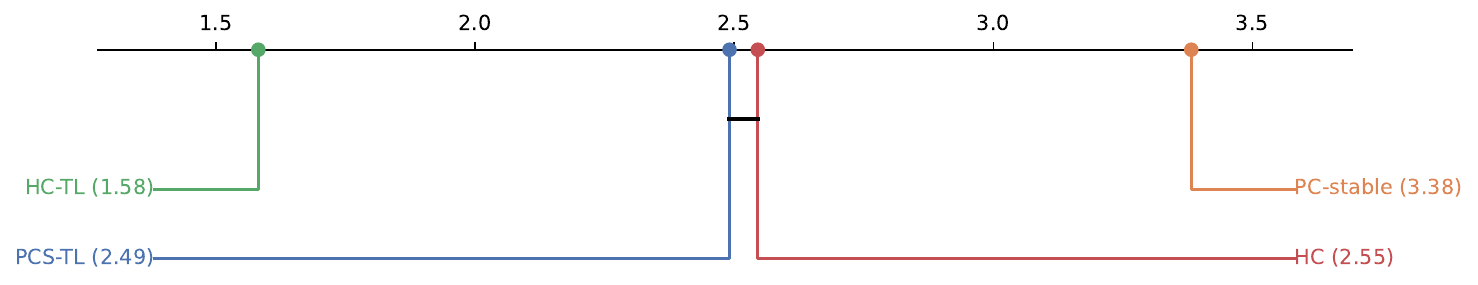}
        \label{fig:logl_010p_cd}
    }    
    \subfigure[Three auxiliary sources with $5\%$, $10\%$ and $20\%$ of the arcs modified]{
        \includegraphics[width=0.9\linewidth]{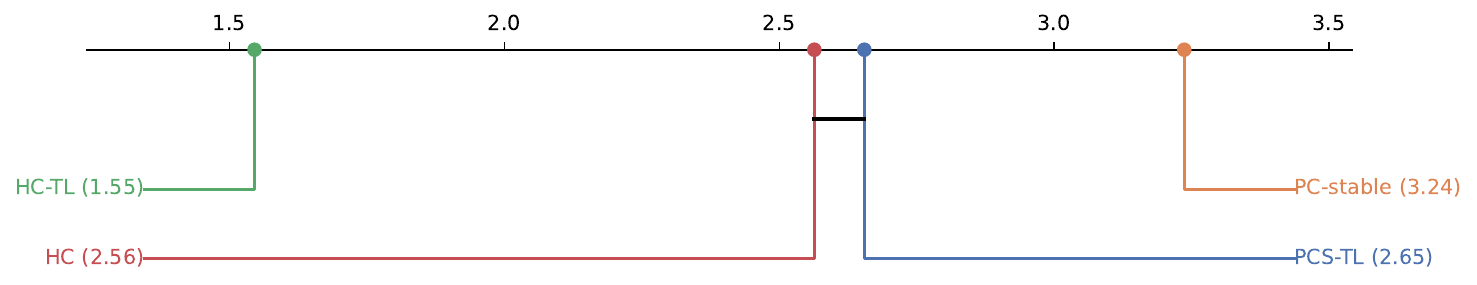}
        \label{fig:logl_51020p_cd}
    }
    \caption{Critical difference diagram for the network's log-likelihood results. Evaluation for less than $525$ target instances.}
    \label{fig:logl_cd}
\end{figure}

\begin{figure}[!ht]
    \centering
    \subfigure[Two auxiliary sources with $0\%$ and $10\%$ of the arcs modified]{
        \includegraphics[width=0.44\linewidth]{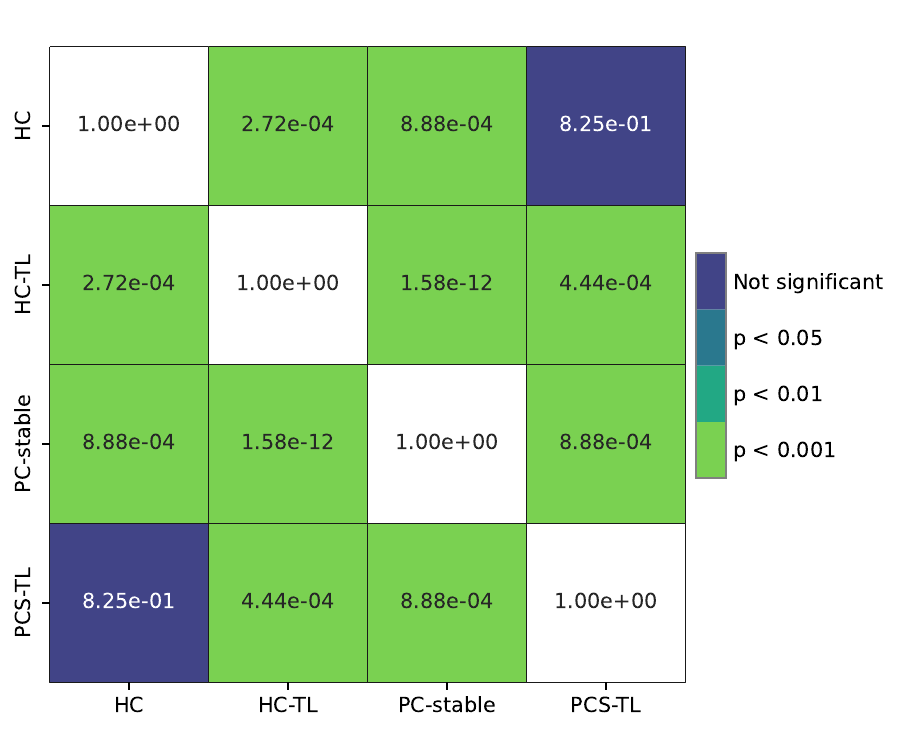}
        \label{fig:logl_010p_heat}
    }    
    \subfigure[Three auxiliary sources with $5\%$, $10\%$ and $20\%$ of the arcs modified]{
        \includegraphics[width=0.44\linewidth]{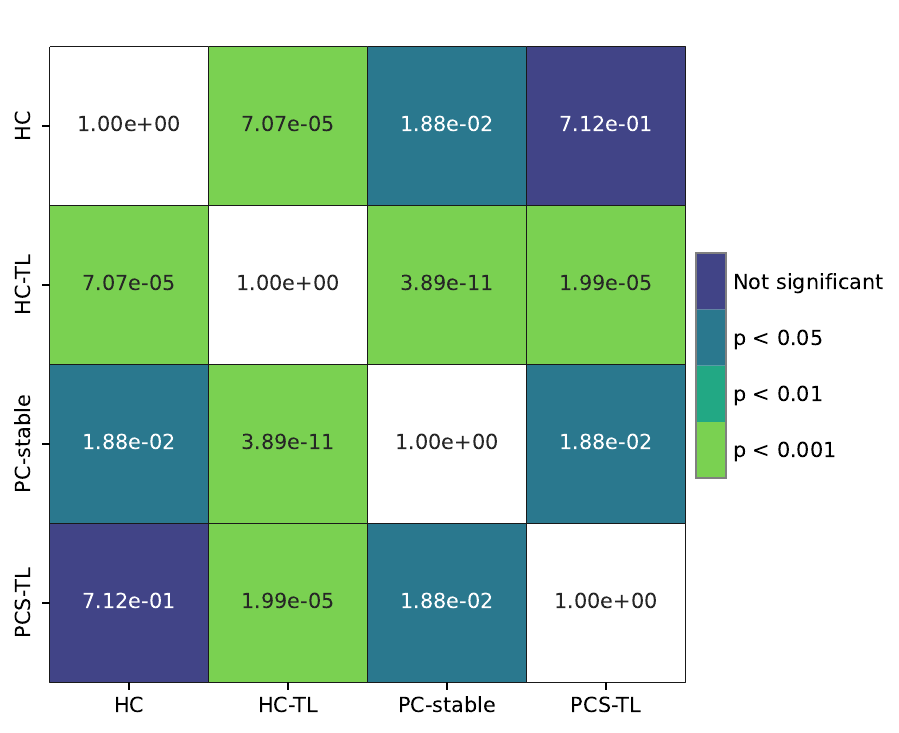}
        \label{fig:logl_51020p_heat}
    }
    \caption{
     Significance heatmap with $p$-values for the network's log-likelihood results. Evaluation for less than $525$ target instances.}
     
    \label{fig:logl_heat}
\end{figure}

In previous subsections, we discussed the log-likelihood and DHD results achieved by the different models. Now, we will analyze whether those comments correspond to a real statistical difference between them. As we exposed at the beginning of this section, we will perform a Friedman test with a Bergmann-Hommel \textit{post-hoc} analysis. This study joins all the previous results for the small synthetic SPBNs, the medium and large synthetic GBNs, and the UCI datasets. However, to study how negative transfer affects the results statistically, we analyzed the performance of the models for the two and three auxiliary sources settings separately (see (a) and (b) for 
Figures \ref{fig:DHD_cd}, \ref{fig:DHD_heat}, \ref{fig:logl_cd}, and \ref{fig:logl_heat}). The previous experiments show a convergence of the transfer learning methods with their counterparts (PC-stable with PCS-TL and HC with HC-TL) for all domains by the time 500 target instances have been reached. Therefore, we performed the analysis for fewer than 525 target instances. The results are presented in a critical difference diagram \citep{critical_diff}. Here, the horizontal lines connect models without significant differences, sorted from left to right according to their mean rank (the number in parentheses). The better the model, the lower the number. 

Figure \ref{fig:DHD_cd} presents the critical difference diagram of the DHD results for each method, and Figure\ref{fig:DHD_heat} shows the pairwise $p$-values with the critical differences in a heatmap. Note that there are no significant differences between the transfer learning structures for these amounts of instances, but there are with their respective counterparts. This is particularly evident for two auxiliary sources, Figure \ref{fig:DHD_010p_cd}. For three auxiliary sources, Figure \ref{fig:DHD_51020p_cd}, there is a clear worsening of the PCS-TL algorithm. The mean rank of the algorithm decreased by $0.5$ points and is connected to HC with a horizontal line. The $p$-value between these two, PCS-TL and HC, went from $p=1.06 \times 10^{-2}$ for two auxiliary sources (Figure \ref{fig:DHD_010p_heat}) to not significant for three (Figure \ref{fig:DHD_51020p_heat}).
This may be due to the sources with a greater presence of incorrect arcs, as is the case for the source with $20\%$ of the arcs modified. In contrast,
The mean rank of HC-TL barely changes under the same circumstances, which indicates the positive effect of the risk function (Equation \eqref{eq:risk}) during the learning process. This robustness against noise is also verified by looking at the $p$-values between HC-TL and HC, for two ($p=1.06 \times 10^{-2}$) and three ($p=4.01 \times 10^{-2}$) auxiliary sources, respectively.

For the log-likelihood results in Figures \ref{fig:logl_cd} and \ref{fig:logl_heat}, the same conclusion can be derived. The deterioration of PCS-TL for three sources is reflected in the mean rank. In both settings, there is a line connecting the PCS-TL and HC models. However,  the mean rank of PCS-TL is slightly higher in Figure \ref{fig:logl_010p_cd} than it is in Figure \ref{fig:logl_51020p_cd}, where the ranks of the other algorithms remain the same. This is reflected in the $p$-value between PCS-TL and HC-TL ($p=1.99 \times 10^{-5}$), which is smaller than it is for HC and HC-TL ($p=7.07 \times 10^{-5}$) in Figure \ref{fig:logl_51020p_heat}, indicating a higher distance with respect to PCS-TL than HC.
In any case, both transfer learning methods resulted in significantly better models than their counterparts according to the log-likelihood results. It is worth mentioning that the differences in log-likelihoods for the constraint-based and score-based algorithms correspond to their learning approaches. Score-based models look for the structure that maximizes the log-likelihood, while constraint-based models evaluate the conditional independence. For instance, in Figure \ref{fig:DHD_010p_cd}, PCS-TL achieved a higher mean rank than HC-TL, despite the not significant differences, while in Figure \ref{fig:logl_010p_cd} HC-TL presents significantly better results.

\section{Conclusion}
\label{sec:conclusion}
This work has overcome the problem of using transfer learning techniques for nonparametric Bayesian networks with scarce data, a field for which there is an evident lack of literature. To this end, we proposed two structure transfer learning methods for PC-stable and HC, PCS-TL and HC-TL, and a log-linear pooling for the fusion of parameters. For HC-TL, we proposed the $S_{\text{CVTL}}^k$ score and the risk metric $R(\bm{\theta}_i^{T}, \bm{\Theta}_i^{S})$, while for PCS-TL, we defined the transfer learning $p$-value $P_{\text{TL}}(X,Y|\textbf{Z})$ to combine the opinion of multiple nonparametric CI tests.  In both scenarios, we leverage the source’s opinion and propose ways of reducing the negative transfer problem. In addition, the use of the SJS divergence and the target trust factor $\eta$, helped to evaluate the relatedness between domains and reduce the dependency of the sources when the target's data increases.

We analyzed the behavior of these proposals for synthetic datasets with Gaussian and nonparametric distributions with 7 to 64 variables, and datasets from the UCI repository with 7 to 24 variables. In all these scenarios, we evaluated the performance of our methods with two and three auxiliary sources, adding noise and modifications to test the resilience to negative transfer. We repeated the experiments several times, showing the average results and demonstrating consistency throughout the process. Finally, we showed real statistical proof of the enhanced performance of our methods before the convergence of the target task alone with the transfer learning task. Our two proposals demonstrated their capability of improving the learning performance of nonparametric Bayesian networks with scarce data. Consequently, we provide a reliable way of reducing the time needed for deploying robust nonparametric Bayesian networks in real industrial environments.

In future works, we will test our methods in real industrial environments, and future research directions could be focused on reducing the effect of negative transfer in the PCS-TL algorithm.

\section*{Acknowledgments}
This work was partially supported by the Ministry of Science, Innovation and Universities under Project AEI/10.13039/501100011033-PID2022-139977NB-I00, Project PLEC2023-010252/MIG-20232016 and DIN2024-013310 (Doctorados Industriales). Also, by the Autonomous Region of Madrid under Project ELLIS Unit Madrid and TEC-2024/COM-89.

\section*{Data availability}
All our synthetic functions and research data are publicly available at: \url{https://github.com/rafasj13/TransferPCHC}.  

\bibliographystyle{unsrt}  


\appendix
\section{Synthetic SPBNs}
\label{apendix:synthetic_data}

Synthetic SPBN 1:
\begin{align}
    &f(a) \sim \mathcal{N}(\mu_A = 3, \sigma_A = 2) \nonumber \\
    &f(b|a) \sim \mathcal{N}(\mu_B = a \cdot 0.5, \sigma_B = 2) \nonumber \\
    &f(c|a) \sim 0.45 \cdot \mathcal{N}(\mu_{C_1} = a \cdot 0.5, \sigma_{C_1} = 1.5) + 0.55 \cdot \mathcal{N}(\mu_{C_2} = 5, \sigma_{C_2} = 1) \nonumber \\
    &f(d|b, c) \sim 0.5 \cdot \mathcal{N}(\mu_{D_1} = c \cdot b \cdot 0.5, \sigma_{D_1} = 1) + 0.5 \cdot \mathcal{N}(\mu_{D_2} = 3.5, \sigma_{D_2} = 1) \\
    &f(e|d, c) \sim 0.5 \cdot \mathcal{N}(\mu_{E_1} = d + c, \sigma_{E_1} = 1) + 0.5 \cdot \mathcal{N}(\mu_{E_2} = 2, \sigma_{E_2} = 1) \nonumber \\
    &f(f|e, d, a) \sim 0.5 \cdot \mathcal{N}(\mu_{F_1} = e + d, \sigma_{F_1} = 1) + 0.5 \cdot \mathcal{N}(\mu_{F_2} = 0.7\cdot a, \sigma_{F_2} = 0.5) \nonumber \\
    &f(g|c) \sim \mathcal{N}(\mu_{G} = c \cdot 0.3, \sigma_{G} = 2) \nonumber
\end{align}

Synthetic SPBN 2:
\begin{align}
    &f(a) \sim \mathcal{N}(\mu_A = 4, \sigma_A = 1.5) \nonumber \\
    &f(b|a) \sim 0.4 \cdot \mathcal{N}(\mu_{B_1} = a \cdot 1.2, \sigma_{B_1} = 1.1) + 0.6 \cdot \mathcal{N}(\mu_{B_2} = 1, \sigma_{B_2} = 1) \nonumber \\
    &f(c|a) \sim 0.5 \cdot \mathcal{N}(\mu_{C_1} = a + 1, \sigma_{C_1} = 1.2) + 0.5 \cdot \mathcal{N}(\mu_{C_2} = 1, \sigma_{C_2} = 1) \nonumber \\
    &f(d|a) \sim \mathcal{N}(\mu_D = a \cdot 0.8, \sigma_D = 1.3) \nonumber \\
    &f(e|c) \sim 0.6 \cdot \mathcal{N}(\mu_{E_1} = c \cdot 1.2, \sigma_{E_1} = 1.3) + 0.4 \cdot \mathcal{N}(\mu_{E_2} = -1, \sigma_{E_2} = 1.5) \nonumber \\
    &f(h|d) \sim 0.6 \cdot \mathcal{N}(\mu_{H_1} = d \cdot 2, \sigma_{H_1} = 1.2) + 0.4 \cdot \mathcal{N}(\mu_{H_2} = 0, \sigma_{H_2} = 1.8) \nonumber \\
    &f(i|b) \sim \mathcal{N}(\mu_I = b \cdot 0.6, \sigma_I = 2)  \\
    &f(j|e) \sim \mathcal{N}(\mu_J = e \cdot 0.7, \sigma_J = 1.7) \nonumber \\
    &f(f|c, h) \sim 0.5 \cdot \mathcal{N}(\mu_{F_1} = c \cdot 1.1 + h, \sigma_{F_1} = 1) + 0.5 \cdot \mathcal{N}(\mu_{F_2} = 15, \sigma_{F_2} = 1.2) \nonumber \\
    &f(g|d, j) \sim 0.5 \cdot \mathcal{N}(\mu_{G_1} = d \cdot 0.8 + j, \sigma_{G_1} = 1) + 0.5 \cdot \mathcal{N}(\mu_{G_2} = 0, \sigma_{G_2} = 1) \nonumber \\
    &f(k|f) \sim \mathcal{N}(\mu_K = f \cdot 0.3, \sigma_K = 2) \nonumber \\
    &f(l|a, c, f, h, d) \sim 0.5 \cdot \mathcal{N}(\mu_{L_1} = a + c + f, \sigma_{L_1} = 1) + 0.5 \cdot \mathcal{N}(\mu_{L_2} = h \cdot 0.6 + d, \sigma_{L_2} = 1.5) \nonumber \\
    &f(m|b, e, g, j) \sim 0.4 \cdot \mathcal{N}(\mu_{M_1} = b + e + g, \sigma_{M_1} = 1.2) + 0.6 \cdot \mathcal{N}(\mu_{M_2} = j \cdot 0.7, \sigma_{M_2} = 1.3) \nonumber 
\end{align}

Synthetic SPBN 3:
\begin{align}
    &f(a) \sim 0.5 \cdot \mathcal{N}(\mu_{A_1} = 4, \sigma_{A_1} = 2) + 0.5 \cdot \mathcal{N}(\mu_{A_2} = 1, \sigma_{A_2} = 1) \nonumber \\ \nonumber
    &f(b|a) \sim \mathcal{N}(\mu_B = a \cdot 0.5, \sigma_B = 2) \\ \nonumber
    &f(c|b) \sim \mathcal{N}(\mu_C = b \cdot 2, \sigma_C = 1.5) \\ \nonumber
    &f(d|b) \sim 0.5 \cdot \mathcal{N}(\mu_{D_1} = b - 1, \sigma_{D_1} = 1) + 0.5 \cdot \mathcal{N}(\mu_{D_2} = 10, \sigma_{D_2} = 1.5) \\ 
    &f(e|d) \sim 0.5 \cdot \mathcal{N}(\mu_{E_1} = d \cdot 2, \sigma_{E_1} = 1.5) + 0.5 \cdot \mathcal{N}(\mu_{E_2} = 3, \sigma_{E_2} = 1) \\ \nonumber
    &f(f|d) \sim 0.6 \cdot \mathcal{N}(\mu_{F_1} = d \cdot 1.5, \sigma_{F_1} = 1.5) + 0.4 \cdot \mathcal{N}(\mu_{F_2} = 0, \sigma_{F_2} = 1) \\ \nonumber
    &f(g|c) \sim \mathcal{N}(\mu_G = c \cdot 0.3 + 5, \sigma_G = 1) \\ \nonumber
   &f(h|c) \sim 0.5 \cdot \mathcal{N}(\mu_{H_1} = c \cdot 0.5, \sigma_{H_1} = 1) + 0.5 \cdot \mathcal{N}(\mu_{H_2} = 10, \sigma_{H_2} = 1) 
\end{align}

Synthetic SPBN 4:
\begin{align}
    &f(a) \sim \mathcal{N}(\mu_A = 5, \sigma_A = 2) \nonumber \\
    &f(b|a) \sim \mathcal{N}(\mu_B = a +2, \sigma_B = 1.5) \nonumber \\
    &f(c|a) \sim 0.4 \cdot \mathcal{N}(\mu_{C_1} = a + 2, \sigma_{C_1} = 1) + 0.6 \cdot \mathcal{N}(\mu_{C_2} = 1, \sigma_{C_2} = 1.5) \nonumber \\
    &f(d|b) \sim 0.5 \cdot \mathcal{N}(\mu_{D_1} = b \cdot 0.8, \sigma_{D_1} = 1.5) + 0.5 \cdot \mathcal{N}(\mu_{D_2} = 15, \sigma_{D_2} = 1.5) \nonumber \\
    &f(e|c) \sim \mathcal{N}(\mu_E = c \cdot 0.7, \sigma_E = 2) \nonumber \\
    &f(f|c) \sim 0.5 \cdot \mathcal{N}(\mu_{F_1} = c \cdot 1.2, \sigma_{F_1} = 1.5) + 0.5 \cdot \mathcal{N}(\mu_{F_2} = -3, \sigma_{F_2} = 1) \nonumber \\
    &f(g|d) \sim 0.6 \cdot \mathcal{N}(\mu_{G_1} = d + 4, \sigma_{G_1} = 1) + 0.4 \cdot \mathcal{N}(\mu_{G_2} = 8, \sigma_{G_2} = 1.5) \nonumber \\
    &f(h|d) \sim \mathcal{N}(\mu_H = d \cdot 0.4, \sigma_H = 2)  \\
    &f(k|d) \sim \mathcal{N}(\mu_K = d \cdot 0.5, \sigma_K = 2.5) \nonumber \\
    &f(i|e) \sim 0.55 \cdot \mathcal{N}(\mu_{I_1} = e \cdot 1.3, \sigma_{I_1} = 2) + 0.45 \cdot \mathcal{N}(\mu_{I_2} = 0, \sigma_{I_2} = 1) \nonumber \\
    &f(j|e) \sim \mathcal{N}(\mu_J = e \cdot 0.5, \sigma_J = 2) \nonumber \\
    &f(o|f) \sim 0.3 \cdot \mathcal{N}(\mu_{O_1} = f + 1, \sigma_{O_1} = 1.4) + 0.7 \cdot \mathcal{N}(\mu_{O_2} = -2, \sigma_{O_2} = 0.7) \nonumber \\
    &f(m|j) \sim 0.6 \cdot \mathcal{N}(\mu_{M_1} = j \cdot 1.5, \sigma_{M_1} = 1) + 0.4 \cdot \mathcal{N}(\mu_{M_2} = 7, \sigma_{M_2} = 1.5) \nonumber \\
    &f(n|j) \sim 0.4 \cdot \mathcal{N}(\mu_{N_1} = j \cdot 1.1, \sigma_{N_1} = 1.2) + 0.6 \cdot \mathcal{N}(\mu_{N_2} = -1, \sigma_{N_2} = 1.3) \nonumber \\
    &f(l|h) \sim 0.5 \cdot \mathcal{N}(\mu_{L_1} = h \cdot 0.3, \sigma_{L_1} = 1.1) + 0.5 \cdot \mathcal{N}(\mu_{L_2} = 5, \sigma_{L_2} = 1.4) \nonumber
\end{align}

\end{document}